\documentclass[conference,compsoc]{IEEEtran}
\usepackage{tikz}
\usepackage{amsmath}
\usepackage{colortbl} 
\usepackage[ruled]{algorithm2e}
\usepackage{multirow}
\usepackage{subfigure}
\usepackage{enumitem}
\usepackage{soul,color}
\usepackage{filecontents}
\usepackage{mdframed}
\usepackage{url}
\usepackage{pifont}
\usepackage{caption}
\newcommand{\name}{\texttt{TruVRF}\xspace}

\ifCLASSOPTIONcompsoc
  \usepackage[nocompress]{cite}
\else
  \usepackage{cite}
\fi
\ifCLASSINFOpdf
\else
\fi
\begin{document}
%

\title{\texttt{TruVRF}: \underline{T}owards T\underline{r}iple-Gran\underline{u}larity \underline{V}e\underline{r}i\underline{f}ication on Machine Unlearning}


\author{
   \IEEEauthorblockN{Chunyi Zhou\IEEEauthorrefmark{1}\IEEEauthorrefmark{2}, \\ Anmin Fu\IEEEauthorrefmark{2}\IEEEauthorrefmark{3}, Zhiyang Dai\IEEEauthorrefmark{2}}
  
  \IEEEauthorblockA{\textit{\IEEEauthorrefmark{1}School of Computer Science and Technology, Zhejiang University}, China 
  }
  \IEEEauthorblockA{\textit{\IEEEauthorrefmark{2}School of Computer Science and Engineering, Nanjing University of Science and Technology}, China 
  }

  \IEEEauthorblockA{zhouchunyi@zju.edu.cn; \{fuam;dzy\}@njust.edu.cn}
\IEEEauthorblockA{\textit{\IEEEauthorrefmark{3}Corresponding author}}
}


%


\maketitle

\begin{abstract}
The notion of the right to be forgotten has sparked increasing interest in machine unlearning. Nevertheless, a notable challenge remains due to the absence of dependable measurement methods for validating the unlearning process carried out by model providers. This insufficiency presents an opening for unscrupulous model providers to deceive data contributors seeking unlearning. The known plausible approach relies on invasive operation, particularly backdoor injection. However, it poses security concerns and is also inapplicable to legacy data---already released data. To tackle this challenge, this work initializes the first non-invasive unlearning verification framework which operates at triple-granularity (class-, volume-, sample-level) to assess the data facticity and volume integrity of machine unlearning. In this paper, we propose a framework, named \name, encompasses three Unlearning-Metrics, each tailored to counter different types of dishonest model providers or servers (Neglecting Server, Lazy Server, Deceiving Server). Specifically, Unlearning-Metric-I verifies the alignment between the removed class by the server and the contributor's unlearned request. Unlearning-Metric-II measures the quantity of unlearned data (sample count) as per the contributor's request. Unlearning-Metric-III validates the correspondence of a specific unlearned sample with the requested deletion. We conducted extensive evaluations of \name efficacy across three datasets, and notably, we also evaluated the effectiveness and computational overhead of \name in real-world applications for the face recognition dataset. Our experimental results demonstrate that \name achieves robust verification performance: Unlearning-Metric-I and -III achieve over 90\% verification accuracy on average against dishonest servers, while Unlearning-Metric-II maintains an inference deviation within 4.8\% to 8.2\%. Additionally, \name demonstrates generalizability across diverse conditions, including varying numbers of unlearned classes and sample volumes. Significantly, \name is applied to two state-of-the-art unlearning frameworks: SISA \cite{Bourtoule2021} (presented at Oakland'21) and Amnesiac Unlearning \cite{GravesNG21}, representing exact and approximate unlearning methods, respectively, which affirm \name's practicality.
\end{abstract}


%
\IEEEpeerreviewmaketitle

\section{Introduction} \label{sec:introduction}
Recent privacy regulations, such as the European Union's General Data Protection Regulation (GDPR) \cite{GDPR}, California Privacy Rights Act (CPRA) \cite{cpra}, and Personal Information Protection and Electronic Documents Act (PIPEDA) \cite{PIPEDA}, empower users with the right to delete their private data from entities storing it, known as the \textit{right-to-be-forgotten}. In the context of Machine Learning (ML), this right necessitates ML model providers to erase any trace of data requested to be forgotten from the model \cite{XueMCAR16}, driving research into the field of eliminating the impact of data on ML models, termed \textit{machine unlearning}. In recent years, machine unlearning has garnered increasing attention from both academia and industry \cite{CaoYASMY18,DuCLOS19,GinartGVZ19,GolatkarAS20,GuoGHM20,IzzoSCZ21,Neel0S21}.
%

Intuitively, a straightforward approach to machine unlearning involves retraining the entire ML model from scratch using a dataset where the samples to be forgotten are excluded. However, this method involves unacceptable computational overhead, particularly for large-scale datasets, as highlighted~\cite{MarchantRA22,MehtaPSR22,SchelterGD21}, which is difficult to use. Existing frameworks that aim to overcome such a challenge can be categorized into two main categories: exact unlearning and approximate unlearning.
Exact unlearning divides a given dataset into multiple non-overlapped blocks and retrains the block(s) with forgotten sample(s) deleted, thus reducing the computation overhead significantly. Approximate unlearning modifies model parameters to conceal the impact of forgotten samples, eliminating the difference in model parameters between an unlearned model and a model that simply does not learn the forgotten samples.

When considering Machine Learning as a Service (MLaaS), many model providers face challenges in acquiring a sufficient amount of high-quality data internally. As a result, they often rely on outsourcing data from data providers such as Amazon Data Exchange~\cite{AWS}, Kaggle~\cite{kaggle}. A data provider acts as an intermediary between different model providers and many data contributors. The same contributor's data might be provided to different model providers by the data provider. To ensure compliance with \textit{the forgotten right}, the data provider is required to provide transparent privacy policies and usage terms that clearly define the control of the data contributors on their personal data when they supply their data to the data provider. 
In particular, the contractual agreement between the data provider and data contributors should clearly outline the obligations and responsibilities with respect to \textit{the forgotten right}. After data contributors supply their data to the data provider, a data contributor still has the right to request data removal from any models that train on such requested data. In addition, the contractual agreement between data provider and model provider requires that the model provider gives a trained model to the data provider for transparency and facilitating later auditing or regulation. 

\begin{figure}[h]
  \centering
  \includegraphics[width=0.8\linewidth]{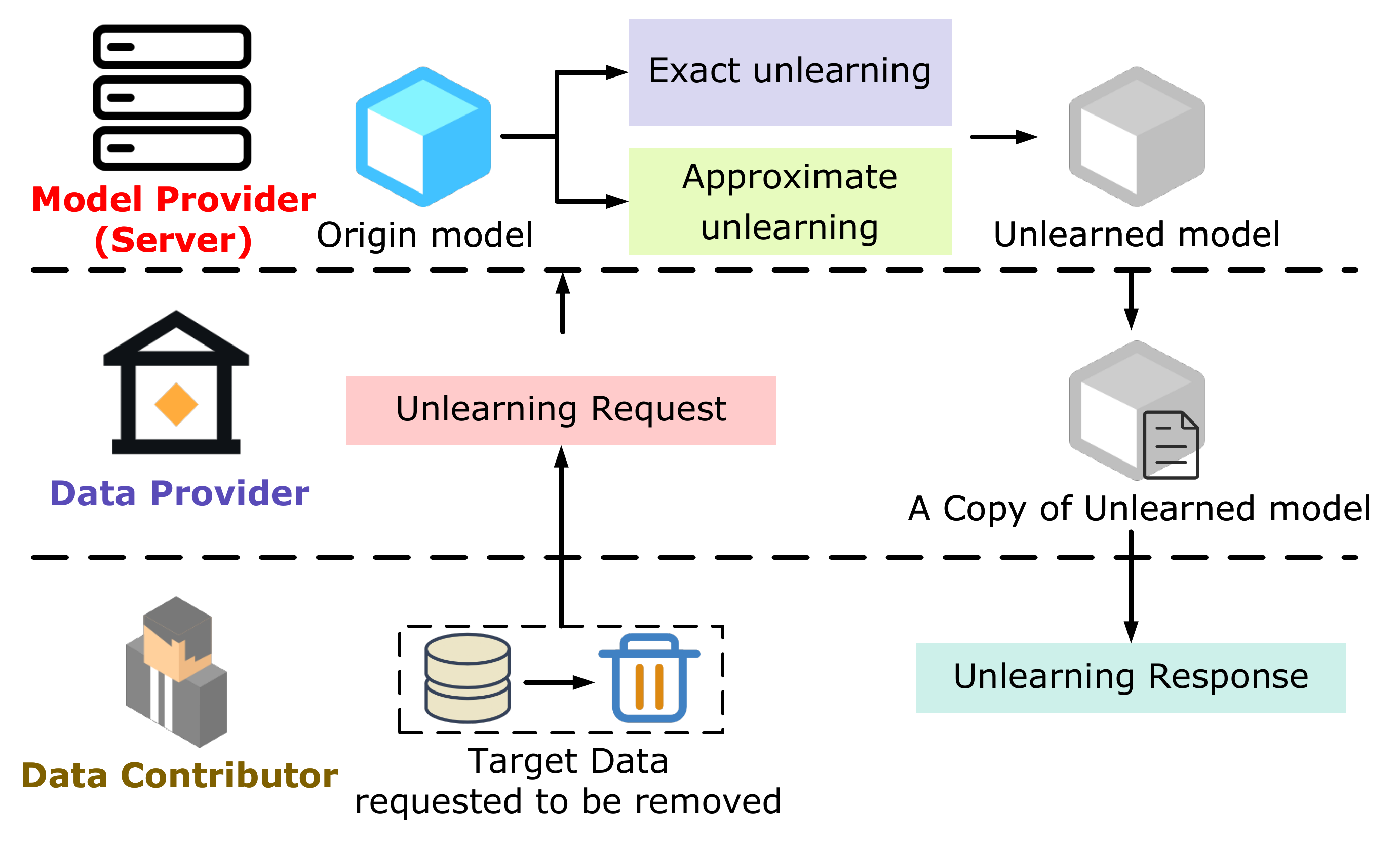}
  \caption{Machine unlearning workflow.}
  \captionsetup{justification=centering}
  \label{Fig1:workflow}
\end{figure}

The workflow of machine unlearning under such an MLaaS is shown in Figure \ref{Fig1:workflow}. Whenever a data contributor needs to delete his/her data, he/she sends the request to the data provider. The data provider then asks all model providers who have used this contributor's requested data to perform machine unlearning on their models. The model provider returns the unlearned model to the data provider once the model unlearning has been done.


However, current machine unlearning frameworks \textit{face the fundamental challenge of lacking a means of verifying the unlearning process}. The data provider has no other option but to trust that the model provider has honestly forgotten the data as requested. 
%
The absence of a verification mechanism renders hesitation or outright refusal by data contributors to share their data. Once data is contributed, contributors lose control over it, exposing their data to privacy risks that cannot be reversed. Specifically, the model provider may act semi-honestly, retaining interest in the data requested by the contributor while avoiding its removal from the model. In some cases, the model provider may simply disregard the data contributor's request to forget the data, or opt to partially unlearn or even unlearn unrelated data to deceive the data contributor. This behavior is rational in practical scenarios, especially when the data contributor seeks to withdraw valuable data crucial to the model provider's interests. In this context, we identify three types of dishonest model providers (hereinafter referred to as ``the server"):

\begin{itemize}[leftmargin=*]
\item {\verb|Neglecting Server|}: The server simply ignores the unlearning request of the data provider and retains the model intact without any operation.
\item{\verb|Lazy Server|}: The server executes the unlearning request of the data provider partially. That is, only selectively forgetting part of the contributor's data.
\item{\verb|Deceiving Server|}: The server executes the unlearning request of the data provider, but forgets other irrelevant samples of the same class and volume as the data provider's unlearning request, aiming to retain the data contributor's target data within the model.
\end{itemize}

The above dishonest unlearning server attempts to violate either \textit{Unlearning Data Facticity} or \textit{Unlearning Volume Integrity} that an honest server should fulfill:
\begin{itemize}[leftmargin=*]
\item {\verb|Unlearning Data Facticity|}: The server should unlearn the data specified in the unlearning request.
\item{\verb|Unlearning Volume Integrity|}: The server should unlearn the same amount of data as the target data in the unlearning request.
\end{itemize}


To this end, we ask the following research questions:

\begin{mdframed}[backgroundcolor=black!10,rightline=false,leftline=false,topline=false,bottomline=false,roundcorner=2mm]
Are there usable verification means for auditing a server's unlearning data facticity and volume integrity? If so, how efficient are they?
\end{mdframed}

\noindent{\textbf{Key Observation}:} During the training phase of ML models, gradient changes play a pivotal role in guiding the convergence process. These changes, indicative of the model's adaptation to the data, are quantifiable through metrics like model sensitivity. Notably, we have observed a correlation between gradient changes and class-specific data volumes within ML models.
In the context of machine unlearning, it is essential to note that the model sensitivity of an unlearned model consistently shifts following the removal of target data. This phenomenon persists under conditions where authentic machine unlearning enforcement is applied, as detailed in Appendix \ref{subsec:exp-Rationale}.

\noindent{\bf Our Solution.} 
Invasive verification methods, such as watermarking based on backdoors \cite{backdoor}, may raise concerns about model security. Additionally, it is inefficient for released data as it is not possible to add watermarks retrospectively if they were not initially added. Based on the key observation, this work attempts the first step towards addressing the above (open) challenging research question by initializing the first holistic and approachable unlearning verification framework, \name, in a non-invasive manner. 
More specifically, we advocate verifying the data facticity and volume integrity of unlearning operations for the sake of unveiling the dishonest behavior by the server. \name entitles the data provider to a viable means of auditing whether the data the contributor wants to withdraw from an ML model (i.e., namely target data in the following descriptions) is forgotten faithfully and completely (i.e., the volume of data) by the server. To ensure verification reliability, we design verification metrics considering three degrees of granularity against various types of dishonest servers: \textit{Unlearned Class Verification}, \textit{Unlearned Volume Verification}, and \textit{Unlearned Sample Verification} from coarse-grained to fine-grained. In the following, unless otherwise stated, we describe \name to verify the exact unlearning. Also \name is applicable to approximate unlearning, as detailed in Section \ref{subsec:exp-Unlearning Verification Generalisability}.


\noindent{\textbf{Methodology}}. Our verification framework is based on the model sensitivity and owns triple levels of granularity: Class Granularity, Volume Granularity, and Sample Granularity, which advocates a responsible auditing approach to the data provider to verify the unlearning behavior by the server. We identify three verification metrics for each granularity resilient to multiple dishonest servers:

\begin{itemize}[leftmargin=*]
\item {\verb|Unlearned Class Verification|}: The data provider extracts the model sensitivity of the origin model $M_o$ and the unlearned model $M_u$. Through contrastive statistical analysis, the data provider can determine the data class unlearned by the server. It could provide the data provider with evidence to verify whether a target class is actually enforced by the server. We denote it as the verification with \textit{Class Granularity}.

\item {\verb|Unlearned Volume Verification|}: The data provider trains multiple shadow models based on the target data that is divided incrementally, and extracts the model sensitivity of each shadow model. In this way, the data provider can learn the volume of unlearned data as a function of model sensitivity change (i.e., we denote it as Unlearning Measurement). Thus, given the Unlearning Measurement, the data provider can infer the volume of unlearned data by the server. We denote it as the verification with \textit{Volume Granularity}.

\item {\verb|Unlearned Sample Verification|}: This metric accomplishes the unlearning data facticity in the level of sample granularity, i.e., the distinguishability between the data contributor-targeted samples and irrelevant samples in the unlearned model $M_u$, which can be used to verify whether the unlearned samples are target samples. The data provider needs to determine the existence of the target sample in unlearned model $M_u$. For the honest model, the model sensitivity extracted by test data and target data is almost the same. But when the server tricks the data provider so that the target samples still exist in the unlearned model $M_u$, these two model sensitivities can produce a distinguishable gap, as described in detail in Section \ref{subsec:exp-Rationale}. Thus, based on this key observation, we propose the Unlearned Sample Verification to verify the existence of the target sample in unlearned model $M_u$. We denote it as the verification with \textit{Sample Granularity}.
\end{itemize}

To this end, we shed light on thwarting the risks of deceiving data contributor and provider, and threatening privacy by the server in the context of machine unlearning, and hereby summarize three types of dishonest servers. 
This work is towarding to a new research line in the context of developing and deploying machine unlearning, so as to ensure the data privacy and realize \textit{the forgotten right}. The main contributions are fourfold:

\begin{itemize}[leftmargin=*]
\item To the best of our knowledge, we are the first that concretely examine the potentially inadvertent misconducts of the server in machine unlearning, and analyze the threats to a data contributor's privacy. Particularly, we identify three types of dishonest servers: Neglecting Server,  Lazy Server, and Deceiving Server.

\item We then propose \name, a machine unlearning verification framework, enabling three levels of granularity based on model sensitivity, which can effectively verify data facticity and volume integrity of machine unlearning, specifically:
    \begin{itemize}
    \item \textit{Unlearned Class Verification} to verify the data class that the server unlearned, which is applicable towards the Neglecting Server scenario.
    \item \textit{Unlearned Volume Verification} to approximately infer the unlearned data volume by the server, which is applicable towards the Lazy Server scenario.
    \item \textit{Unlearned Sample Verification} to judge the existence of the target sample in the unlearned model $M_u$, so as to verify whether the server unlearns the requested sample honestly, which is applicable towards the Deceiving Server scenario.
    \end{itemize}

\item We evaluate \name through extensive experiments on three datasets, CIFAR-10, Fashion-MNIST, and RAF-DB, which demonstrate the verification ability against the three aforementioned types of dishonest servers. Experimental results validate that the unlearning metrics can efficiently verify the unlearned class, volume, and sample. Notably, \name is also validated to be effective and deployable in the real-world  facial recognition application. 

\item We affirm \name's generalisability through the verification evaluations on two state-of-the-art unlearning frameworks: SISA  \cite{Bourtoule2021} (Oakland'21), and Amnesiac Unlearning~\cite{GravesNG21} as representative exact and approximate unlearning, respectively. We further elaborate on the rationale of Unlearning-Metrics based on triple levels of granularity.
\end{itemize}

\noindent{\textbf{Roadmap}}. In Section \ref{sec:preliminary} , we introduce preliminary knowledge about machine unlearning and discuss related work. In Section \ref{sec:scheme}, we define the threat model and then elaborate on the details of \name. We conduct extensive experiments to validate the efficacy of \name in Sections \ref{sec:experiments} and \ref{sec:discussion}. We conclude the paper in Section \ref{sec:conclusion}. 

\begin{table}
	\centering
 \caption{A summary of notations}   
 \captionsetup{justification=centering}
	\label{tab1:Notation}

    \resizebox{1\columnwidth}{!}{
	\begin{tabular}{c|l}   
		\hline  
		\textbf{Notation} & \textbf{Description}  \\   
		\hline  
		$x$   & Target data that the data contributor requests to unlearn.\\  
		\hline  
		$M_{o}$  & Origin ML model before machine unlearning.\\  
		\hline  
		$M_{u}$  &  Unlearned ML model after machine unlearning.\\  
		\hline  
		$\delta$  & Transformation on $M_{o}$ in approximate unlearning.\\  
		\hline  
		$\theta _{o}$ & Model Parameter Vector in $M_{o}$\\
		\hline
		$\theta _{u}$ & Model Parameter Vector in $M_{u}$\\
		\hline
		$\alpha$ & Learning Rate\\
		\hline
		$MS$ & Model Sensitivity\\
		\hline
		$DS$ & Model Sensitivity Difference\\
		\hline
		$UM$ & Unlearning Measurement\\
		\hline
	\end{tabular}}  
\end{table}  

\section{Preliminary and Related Work} \label{sec:preliminary} 
We first introduce machine unlearning, and briefly describe exact unlearning and approximate unlearning. We then discuss existing potential metrics/methods that might be indirectly utilized for unlearning verification.

\subsection{Machine Unlearning Preliminary} \label{subsec:unlearning}
Benefiting from the recent legislation such as GDPR and CCPA, a user can legally request data controllers to delete his/her individual data, which is formalized in the right to be forgotten. In the context of machine unlearning, an ML model has learned the features of user data, and cannot fulfill this right under formal ``deletion''. To this end, an ML owner needs to resort to machine unlearning to erase the impact of the data to be forgotten on the model. There are two categories of machine unlearning: exact unlearning and approximate unlearning. We summarize the notations in Table \ref{tab1:Notation} to ease the following descriptions. 

\subsubsection{Exact Unlearning}
This leads to the exact trace removal of target data (i.e., a volume of samples or just a specific sample) from the training set in the unlearned model. The most straightforward way is to train the ML model from scratch after removing the target data $x$ from the training set $D$. We denote the origin ML model before unlearning as $M_{o}$, and the ML model after unlearning as $M_{u}$. Thus, exact unlearning can be formalized as: 
\begin{align}
M_{u} & = M_{o}\backslash x=\textsf{train} (D\backslash x).
\end{align}
Although training the whole model can perfectly satisfy the goal of machine unlearning, it would incur unbearable computation overhead as the model structure is often complicated or the training set scale is large. Due to the equal right-to-be-forgotten for each user in the multi-user scenario, a model owner is infeasible to satisfy users' unlearning requests in this means of retraining from scratch all the time, even the model provider/server intends to do so. Therefore, practical unlearning frameworks are proposed, such as SISA \cite{Bourtoule2021}, pursuing to mitigate the computational overhead of the whole model retraining. Essentially, these methods retain the benefits of exact unlearning in terms of discarding the origin model, removing the target data from the training set and then updating the unlearned model through a more efficient manner, e.g., SISA only retrains a sub-model on a small sliced data block from which the requested data has been removed to update the ensemble model decision. 

\subsubsection{Approximate Unlearning}
It chooses an alternative path, focusing on how to transform the origin model into another one without the target data effect in the training set, and make the twin models infinitely close in the parameter space. Formally, we can denote it as: $M_{u}$=~$M_o$+$\delta$$\approx$~$M_{o}$$\backslash$$x$, where $\delta$ represents the transformation operation carried out by the model owner on  $M_{o}$, such as adding noise. Approximate unlearning methods are building upon their ``unlearning'' on reproducing similar properties of exact unlearned models. Despite that \name is mainly motivated to address the unlearning verification on exact unlearning, it is also applicable for approximate unlearning.

\subsection{Related Work}

\subsubsection{Exact Unlearning} \label{subsec:exact}
The notion of machine unlearning was first proposed by Cao et al. \cite{Cao2015}, which aims to implement the right to be forgotten and decrease the damage of poison data in the machine learning context. They designed an efficient unlearning approach by transforming learning algorithms (specifically, SVM, naïve Bayes, and decision tree) into a summation form. It introduces a layer of a small number of summations between the learning algorithm and the training data to break down the dependencies. When receiving the unlearning request, the model owner only needs to remove the transformations about the target data from the summations, which reduces the computation cost. However, this method is not suitable for deep learning, such as neural networks.
Cao et al. \cite{CaoYASMY18} proposed a causal unlearning method Karma, which searches through different subsets of training samples and returns the subset that causes the most misclassifications as the set of polluted training samples. By removing the problematic subset, Karma can reduce the impact of data pollution. However, Karma is only applicable for ML classifiers based on SVM and naïve Bayes, while could not be implemented in a more complicated model, such as neural networks. 

Later, the exact unlearning is extended to deep learning. 
Bourtoule et al. \cite{Bourtoule2021} proposed the SISA framework, which is a representative approach to date for exact unlearning. SISA divides the training data into multiple disjoint shards, and trains sub-models based on them respectively. The final model is presented in the form of an aggregation or ensemble of these sub-models. When the unlearning invokes, SISA only needs to retrain the sub-model corresponding to the shard under which the target data falls. Thus, SISA substantially obviates the computation overhead of retraining all the rest sub-models. Following this work, Chen et al. \cite{Chen000H022} proposed GraphEraser, tailored to graph data based on SISA. It adopts the existing unlearning framework to the graph domain. Wu et al. \cite{WuDD20} proposed DeltaGrad for rapid retraining machine learning models, which is based on the information cached during the training phase. Yan et al. \cite{ijcai2022p556} proposed ARCANE. After the selection of training data, it is divided into many shards. ARCANE uses each shard to train a model, and records the historic training state. The model will be traced back to the corresponding state and continue training upon receiving the unlearning request. 

\subsubsection{Approximate Unlearning} \label{subsec:approximate}
Approximate unlearning attempts to unlearning in the perspective of model parameters, i.e., by modifying the model parameters so that the unlearned model is close to the model trained when the training set never contains the target data \cite{Neel0S21,Zhang0ZCL22,GuptaJNRSW21}. Graves et al. \cite{GravesNG21} designed the Amnesiac Machine Learning, which relabels the sensitive data with randomly selected incorrect labels and then retrains the model for some iterations on the modified dataset. 
Additionally, the model owner records the training data, and saves the parameter updates for each training batch. When the unlearning happens, the model undoes the corresponding update. 
Baumhauer et al. \cite{BaumhauerSZ22} developed linear filtration, an approximate unlearning method for the sanitization of classification models that predict logits, after class-wise deletion requests. Guo et al. \cite{GuoGHM20} proposed a Certified Removal Mechanism for linear classifiers to make the unlearned model close to the model that never observed the data at the beginning. Izzo et al. \cite{IzzoSCZ21} introduced projective residual updates to reduce the unlearning time complexity. Ginart et al. \cite{GinartGVZ19} proposed two provably efficient unlearning algorithms based on K-means, which enable efficient data deletion in ML. 
Chundawat et al. \cite{ChundawatTMK23} introduced the concept of zero-shot machine unlearning that caters for the extreme but practical scenario where zero original data samples are available for use. They proposed two methods of error minimizing-maximizing noise and gated knowledge transfer to achieve zero-shot machine unlearning. Golatkar et al. \cite{GolatkarAS20} investigated the connection between Differential Privacy and the stability of SGD, and designed a selective forgetting method in ML. Warnecker et al. \cite{WarneckePWR23} proposed a framework of approximate unlearning to erase the features and labels, which is based on influence functions. It can adapt the influence of training data on a learning model retrospectively. 

Although approximate unlearning can make the unlearned model closer to the exact unlearned model in parameter space, there are still certain limitations. Thudi et al. \cite{ThudiJSP22} utilized the forging technology to prove that approximate unlearning is self-contradicted. It can satisfy the definition of approximate unlearning in parameter space without any modification when the forging match is found. Meanwhile, approximate unlearning would have an impact on the model accuracy \cite{GravesNG21}. When the unlearning request is on a large scale, the model efficacy drops non-negligibly.

\subsubsection{Indirect Unlearning Verification Methods} \label{subsection:verification method} 
Existing indirect unlearning verification methods are ad-hoc, focusing on special attacks and proof of learning technology to verify the unlearning indirectly. Specifically, membership inference attacks \cite{ShokriSSS17,SongS19,ChenCCS2021} are used to verify the effectiveness of approximate unlearning \cite{BaumhauerSZ22,GravesNG21}, which demonstrates that the unlearned model cannot be distinguished from the model that has never trained on the target data. 
Sommer et al. \cite{backdoor} verified the exact unlearning by a backdoor attack~\cite{gao2020backdoor}. A user executes a backdoor injection during model training. When the server does not delete the target data, the predictions of backdoor samples are the target labels with high probability. However, there are several notable limitations to verify unlearning based on backdoor attacks. Firstly, it introduces security concerns through injecting backdoors, which poses the safe usage of the model under insidious risks, especially for other data contributors. Secondly, as acknowledged \cite{backdoor}, the verification accuracy drops once the server adopts the backdoor defense, such as Neural Cleanse~\cite{WangYSLVZZ19}---a model-based detection defense. As machine unlearning usually refers to the data outsourcing scenario, we note that training-based defenses~\cite{li2021anti,wang2022training} (the defender/user rather than the attacker trains the model) are expected to remove the backdoor effect, making the verification fall.
Thirdly,  when the Lazy Server only unlearns partial data, existing work can only verify the existence of the target data but cannot infer the specific data quantity.

Furthermore, Thudi et al. \cite{ThudiJSP22} improved the proof of learning \cite{JiaYCDTCP21} to the proof of unlearning. By checking the reliability of intermediate models in unlearning, they can verify whether the server is honestly executing the unlearning operation. The similarity between the approximate unlearned model and the retrained model is regarded as unlearning metrics, such as ${\ell _2}$ distance \cite{WuDD20} and KL divergence \cite{GolatkarAS20}. Despite these metrics being straightforward, there are limitations that need to train the model from the scratch, and there exist deviations which are caused by randomness and numerical instabilities in floating point operations. To solve these issues, Thudi et al. \cite{ThudiDCP22} further developed ${\ell _2}$ distance as the verification metric towards approximate unlearning by expanding the SGD algorithm. In addition, a surrogate approach to verifying the efficiency of unlearning is to analyze the privacy leakage of the weight distributions in the unlearned model \cite{SekhariAKS21, GuoGHM20}. Typically, the data privacy that the model could release is information-less as it does not exist in the model. However, the aforementioned methods have limitations, such as verification based on a backdoor attack raising security concerns and might be removed \cite{STRIP, WangYSLVZZ19, Tang0TZ21}. 
Moreover, the proof of unlearning is fragile \cite{ShumailovSKZPEA21}. It is difficult to measure the quality of exact unlearning based on ${\ell _2}$ Distance and KL Divergence as unlearning metrics. Except for guiding unlearned model optimization, these metrics are inadequate in revealing the misconducts of dishonest servers from a quantitative standpoint. 
We prefer denoting them as optimization metrics rather than verification metrics.

The aforementioned unlearning verification methods are not directly and delicately designed for unlearning verification, which fails to verify the data facticity and volume integrity purposefully in diverse request requirements of machine unlearning (i.e., from sample level to volume level and class level). They can only qualitatively analyze whether the server has deleted the data (in particular, TRUE or FALSE), but cannot verify quantitatively, such as accurately inferring the unlearned category, quantity, or target sample existence. Consequently, \name initiates the first systematic study on unlearning verification and aims to use a holistic toolkit to confront dishonest servers in the context of machine unlearning.

\section{Unlearning Verification of \name} \label{sec:scheme}
In this section, we first define the threat model and clarify the capabilities of involved entities (i.e., model provider or the server, data provider and data contributor), then present an overview of \name, followed by its implementation details with three degrees of granularity.

\subsection{Threat Model} \label{subsec:threat model}

As depicted in Figure \ref{Fig1:workflow}, it is assumed that a model provider (the server) is often challenged to acquire a sufficient amount of high-quality data internally. Then the server relies on sourcing data from the data provider who collects data from a multitude of data contributors, e.g., through crowd-sourcing. According to data protection regulations (i.e., \textit{the forgotten right}), data contributors can revoke the contribution of their data. The same data might be provided to different model providers by the data provider. Upon a data deletion request from a data contributor to the data provider, the latter reviews it and submits an unlearning request to the server. Then the server will fulfill the unlearning request using the exact or approximate unlearning methods and return a copy of the model to the data provider for transparency and regulation purposes. Note that this does not infringe on the data privacy of the server because the data provider already has these data. Finally, the data provider responds to the data contributor with the unlearning outcome.

\begin{figure*}[h]
  \centering
  \includegraphics[width = 0.65\textwidth]{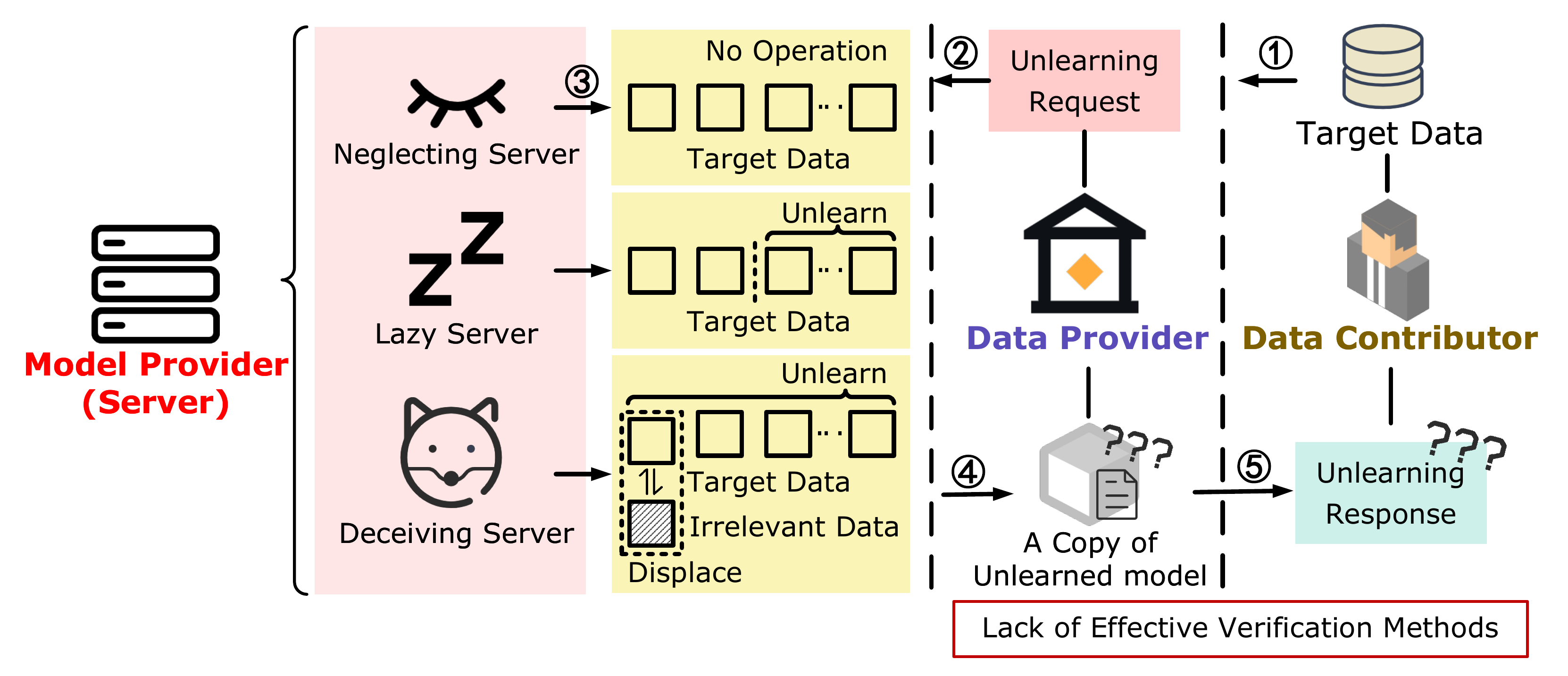}
  \caption{A schematic view of threat model. \textcircled{1} A data contributor requests the data provider to delete the target data from all models that use the target data. \textcircled{2}. The data provider reviews the request of data contributor, and sends unlearning requests to corresponding model providers. \textcircled{3}. A malicious model provider executes dishonest unlearning in three ways. \textcircled{4}. The model provider returns the unlearned model to the data provider after unlearning for auditing. \textcircled{5}. The data provider responds with unlearning result to the data contributor.}
  \label{Fig2:threat model}
\end{figure*}

Specifically, the threat model of machine unlearning is depicted in Figure \ref{Fig2:threat model}. In MLaaS scenarios, there are three entities involved in machine unlearning services: the model provider (the server), the data provider, and the data contributor. We detail the capabilities and knowledge of the above entities as follows:

 \begin{itemize}[leftmargin=*]
  
    \item \textbf{Data Contributor.} The data contributor is the owner of the model training data. Generally, the data contributor would sell the annotated data to the data provider, or complete the crowd-sourcing annotation task published by the data provider. Due to the privacy regulations (i.e., \textit{the forgotten right}) \cite{GDPR}, the data contributor has full control over the data he/she owns. In the context of machine unlearning, the data contributor is allowed to ask the model provider to revoke his/her data contribution to the ML model.
    
    \item \textbf{Data Provider.} The data provider undertakes the tasks of data collection from many data contributors and sells the data to different model providers. When the data provider acquires the data or publishes the crowd-sourcing clickwork, the transparent privacy policies and terms of usage that define data contributors' control over their personal data should be provided clearly. Data provider could access a copy of the model from the model provider who has used the data from the provider. That is, the data provider has access to the origin model $M_o$ and the unlearned model $M_u$ provided by the server to allow later unlearning verification. This model access is factually in alignment with the existing unlearning research \cite{ChenCCS2021,Bourtoule2021,GravesNG21,ThudiJSP22}. In addition, data provider possess substantial computational capabilities and holds some auxiliary data (such as the data contributor's target data and test data) for the purpose of unlearning verification.
    
    \item \textbf{Model Provider (the server).} The server is the model trainer or model owner, with full control of the model. Additionally, the server is responsible for the unlearning operation whenever requested by the data provider. Typically, the server removes the target data by the exact or approximate unlearning method. Then the server would return a copy of the unlearned model to the data provider for transparency purposes. However, the server could be reluctant to honestly perform the unlearning due to incurred cost or model deterioration if the requested data are crucial, where the server deceives the data provider. The server may behave in the following dishonest behaviors: 
    \begin{itemize}
    \item {\verb|Neglecting Server|}: The server simply ignores the  unlearning request and does not perform any operation on the ML model. In this situation, data contributor's target data still remains in the origin model $M_o$, in which the class and data volume will be intact. This type of server violates the Unlearning Data Facticity and Unlearning Volume Integrity.

    \item {\verb|Lazy Server|}: The server executes the data provider's unlearning request partially, by forgetting the target data selectively. In this situation, the server will review the data provider's unlearning request, filter the data class and data volume that the server wants to keep and forget the remaining data he/she is not interested in. This type of server violates the Unlearning Volume Integrity.

    \item {\verb|Deceiving Server|}: The server executes the data provider's unlearning request, but he/she will distort it. Concretely, the server will choose the same class and data volume of other \textit{irrelevant} data to replace the target data. In this situation, the server aims to retain the data of the contributor in the ML model and attempts to deceive the data provider by unlearning other irrelevant data. This type of server violates the Unlearning Data Facticity.
    \end{itemize}
\end{itemize}

\begin{figure*}[h]
  \centering
  \includegraphics[width = 0.8\textwidth]{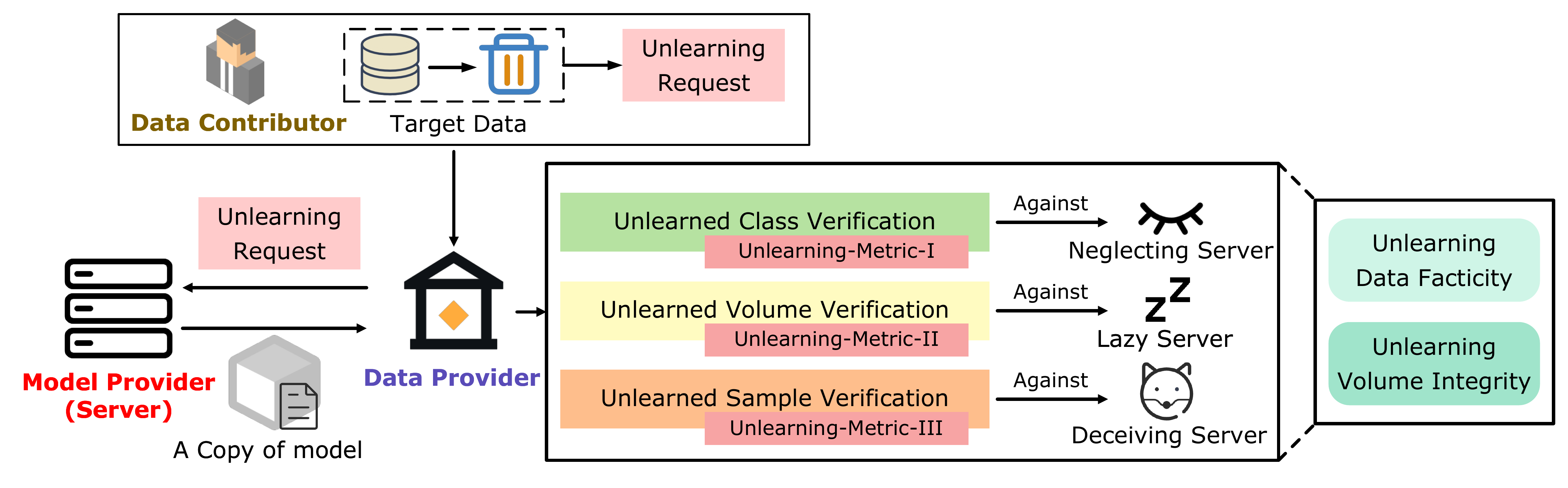}
  \caption{\name overview.}
  \label{Fig3:overview}
\end{figure*}

The proposed \name is to enable the data provider to verify the data facticity and volume integrity of the unlearning performed by the server, i.e., \textbf{whether the server has forgotten the target data} (data facticity) and \textbf{whether the server has forgotten the enough data} (volume integrity). 
The obligation for data providers to furnish unlearning verification services to data contributors stems from the imperative to adhere to prevailing privacy regulations and laws. Incentivized by the prospect of bolstering their credibility, attracting a broader spectrum of data contributors, assuaging privacy apprehensions, and fostering a transparent and robust Data as a Service ecosystem, data providers are compelled to offer such services.
To this end, \name consists of three degrees of verification ability devised upon the key observation of \textit{model sensitivity}.

\subsection{Model Sensitivity} \label{subsec:model sensitivity}


\begin{algorithm}[]  
	\caption{Extracting Model Sensitivity}
	\label{alg1:extract}
	\LinesNumbered 
	\KwIn{\textbf{ ML model $\theta$, auxiliary dataset $D_{\rm{aux}_{c}}$ of class $c$, learning rate $\alpha$}}
	\KwOut{\textbf{Model sensitivity $MS$ of class $c$}}
	Set  $MS=0$\;
		Data provider trains the model $\theta'= \theta. \textsf{train}(D_{{\rm{aux}}_{c}}, \alpha)$\;
		\ForEach{neuron $i$ in $\theta'$}{
		Compute $MS_{c}^{i}=\left | \left (\theta- \theta'  \right )\cdot \frac{1}{{\alpha }} \right |=\left | \frac{\delta }{\delta \theta _{i}}L(\theta ) \right |$\;
		$MS += MS^{i}$\;
		}
\end{algorithm}

The trained ML model learns the feature of training data, so as to classify the test data, which is the inherent characteristic of ML. 
Additionally, we found that the ML model could  memorize the data-distribution characteristic of a training dataset, and inadvertently reflects the distribution in the form of gradient changes. The gradient value is related to the class and volume of training data (detailed in Section \ref{subsec:exp-Rationale}). More precisely, if the sample number of a class in a training dataset is small, the model is unlikely to learn to generalize to that class. Thus, when the model trains on such dataset, it will exhibit a greater gradient effect to change the weights of the corresponding neurons to minimize the expected loss of the model regarding such class. That is, the gradient change is negatively correlated with the number of class samples in the training dataset. We define the gradient change in training as the \textit{model sensitivity}. We note such model sensitivity was recently exploited to profile user preference in the federated learning that leaks the local user's private information~\cite{PPA}. In contrast, we turn this model sensitivity as an asset to address the open challenge of delicately verifying machine unlearning conducted by a dishonest centralized server.

In the context of honest machine unlearning, it retrains the model by removing the requested data from the training set. The model sensitivity is expected to display a notable discrepancy between the origin model $M_o$ and the unlearned model $M_u$ (see experimental validation details in Section \ref{subsec:exp-Rationale}). Attributing to the decrease of training data, the model sensitivity of the requested class will increase, which serves as the key observation for constructing \name. To this end, we quantify the model sensitivity by utilizing the sum of absolute values of the gradient changes before and after the retraining model, which is expressed as:
\begin{equation}
    MS= \sum\left | \frac{\delta }{\delta \theta'}L(\theta ) \right |=\sum\left | \left (\theta- \theta'  \right )\cdot \frac{1}{{\alpha }} \right |,
\end{equation}
where $\theta$ represents the model parameter vector of the origin model, and $\theta'$ represents the model parameter vector of the unlearned model.
\textit{L}($\cdot $) represents the loss function, and $\alpha$ represents the learning rate. The process is detailed in Algorithm \ref{alg1:extract}. In this paper, auxiliary dataset refers to the data used to retrain the ML model by the data provider to extract $MS$, which can be the data contributor's target data or test data. Notably, also as a data curator and aggregator, the data provider has plenty of data.


Then, upon the model sensitivity, we construct three \name metrics serving as verification toolkits against all three types of dishonest servers (Neglecting Server, Lazy Server, and Deceiving Server) to verify the unlearned model $M_u$ returned by them. Figure \ref{Fig3:overview} illustrates the structure of \name, and the detailed Metric design and corresponding verification procedure are presented as follows.

\subsection{Unlearning-Metric-I: Unlearned Class Verification} \label{subsec:Unlearning-Metric-I}

\begin{figure}[htb]
  \centering
  \includegraphics[width = 0.45\textwidth]{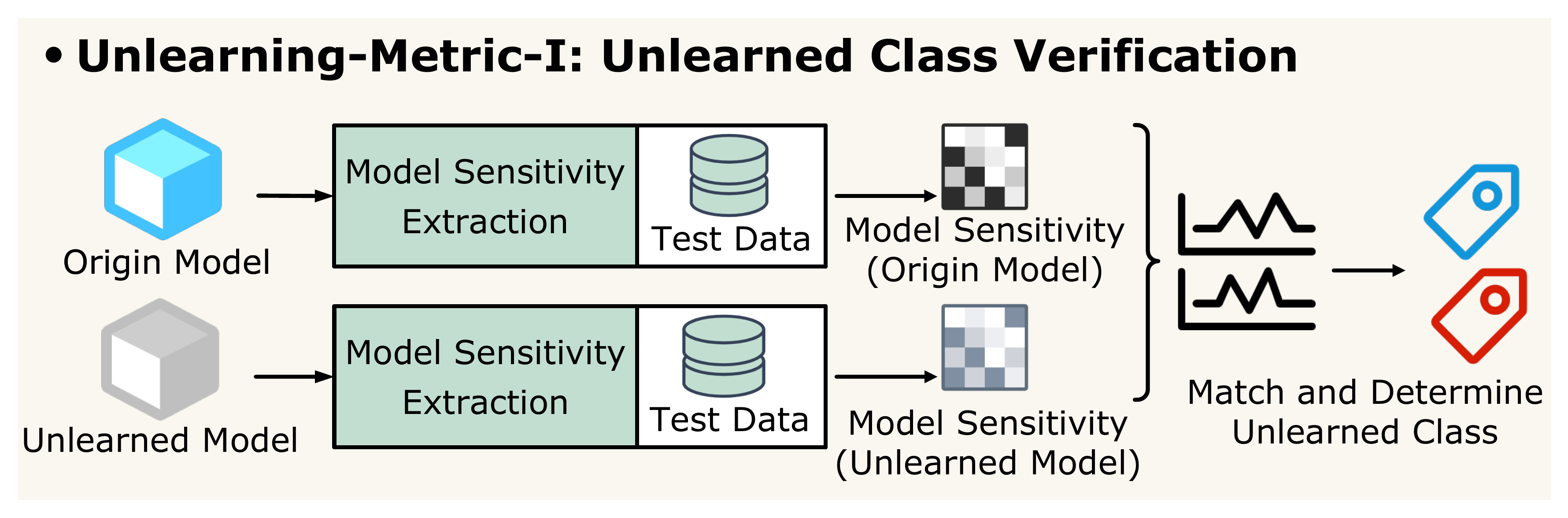}
  \caption{Process of Unlearned Class Verification.}
  \label{Fig4:Unlearning-Metric-I}
\end{figure}

This metric is designed to verify the unlearning result of Class Granularity, and decides which class(es) the server has unlearned. For example, a data contributor wants to withdraw all/some of his/her own face images from a facial recognition model, and we defer to evaluating such a situation in Section \ref{subsec:UMI-face}. After receiving the copy of the unlearned model $M_u$ and the origin model $M_o$ from the server, the data provider starts the verification by continuing training the $M_o$ and $M_u$ upon his/her own testing data. Then the data provider collects the gradient changes during the training and computes the model sensitivity for each target class. We call this process model sensitivity extraction; see Algorithm
\ref{alg1:extract}. All model sensitivity extraction in the following study utilizes this Algorithm \ref{alg1:extract}, and hereinafter referred to as $\textsf{Extract ($\cdot$)}$. The data provider can judge whether the server has forgotten the target class by matching the sensitivity difference of the model for each class in two models. The specific process is shown in Figure~\ref{Fig4:Unlearning-Metric-I}.

If the server has honestly removed the target data and updates the model, the model sensitivity of target class in the unlearned model $M_u$ increases, as detailed in Section \ref{subsubsec:exp-Rationale-Metric-I}. Because the sample number of the corresponding class becomes smaller, the model would exhibit a significant gradient change for this category. Furthermore, the more forgotten, the more obvious the change discrepancy between the origin model $M_o$ and the unlearned model $M_u$ will be. This metric is suitable for verifying Neglecting Server. In this case, the origin model $M_o$ and the unlearned model $M_u$ differ less in the sensitivity of the model, where the verification of the unlearned class could perform well. In other words, Neglecting Server does not remove the target data at all, and training dataset of twin models before and after unlearning has remained identical. 

\noindent\textbf{Rationale of Unlearning-Metric-I.} We evaluate the model sensitivity changes of the Honest Server and the Neglecting Server before and after unlearning respectively. We found that, there is a significant discrepancy of the target class between these two servers, which sets a foundation of designing Unlearned Class Verification. Details are deferred to Section \ref{subsubsec:exp-Rationale-Metric-I}.

\subsection{Unlearning-Metric-II: Unlearned Volume Verification} \label{subsec:Unlearning-Metric-II} 
\begin{figure}[h]
  \centering
  \includegraphics[width = 0.45\textwidth]{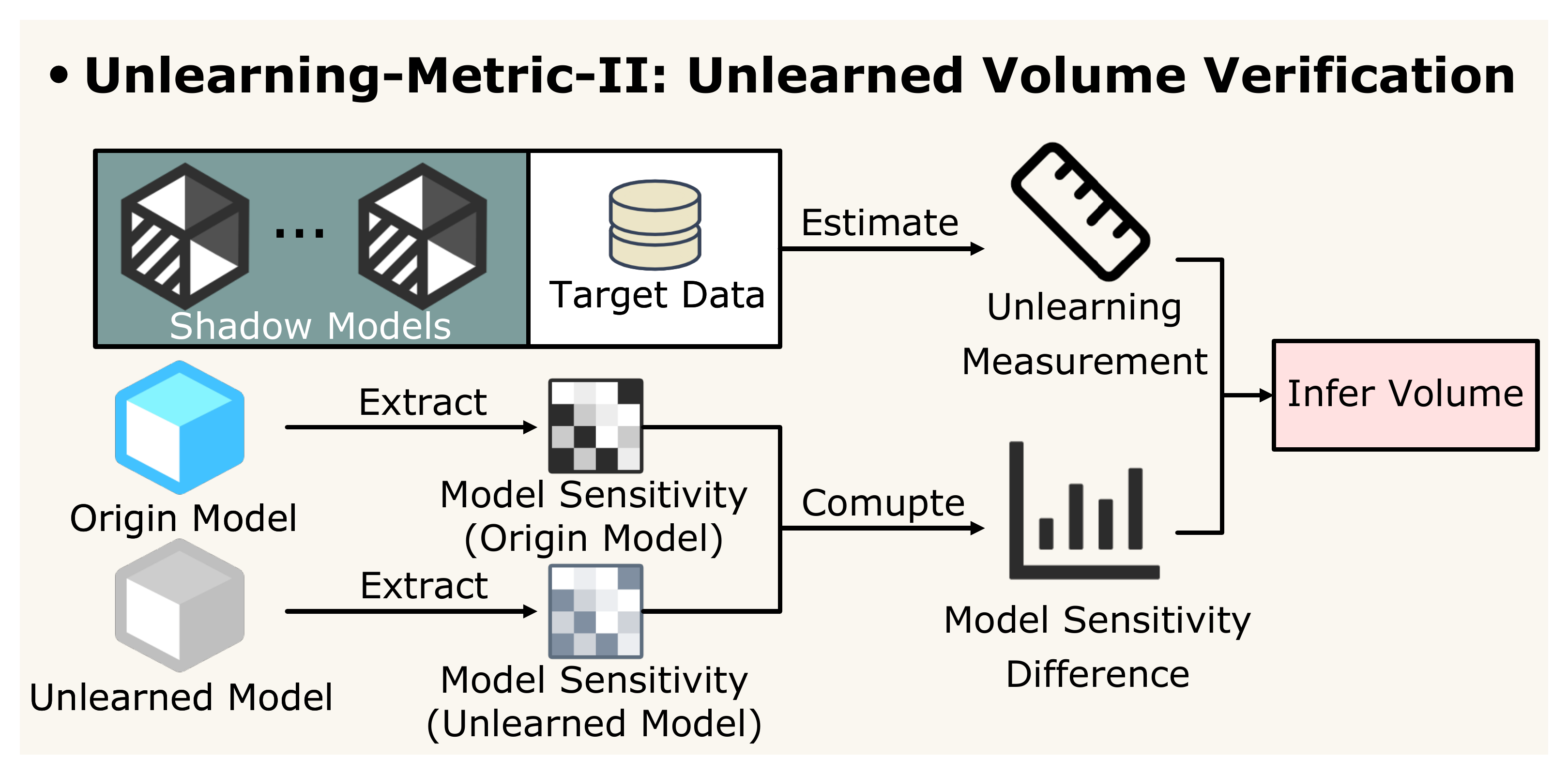}
  \caption{Process of Unlearned Volume Verification.}
  \label{Fig5:Unlearning-Metric-II}
\end{figure}

In more complex scenarios like Lazy Server and Deceiving Server, Unlearning-Metric-I may not fully meet the verification requirements when the data contributor requests the removal of only a portion of a class. In such cases, the server might perform unlearning of the target class data, but not necessarily the requested volume, which can bypass the Class Granularity. Therefore, it becomes necessary to verify the number of unlearned samples in Volume Granularity using statistical methods.

In order to verify whether the server unlearns the requested amount of target data, we design the unlearned volume verification. Firstly, the data provider extracts model sensitivity of target data against both origin and unlearned models according to Algorithm1. Then the data provider computes the model sensitivity difference $DS$, which represents the change of model sensitivity caused by the decrease in the number of samples in this category: 
\begin{equation}
DS=MS_{u}-MS_{o}.
\end{equation}

Metric-II builds on the relationship between the difference in model sensitivity, $DS$, and the volume of requested forgotten data, to decide the number of samples that the server has unlearned. To this end, we define an unlearning measurement, which is to transform from model sensitivity difference to target data volume. As the name indicates, the data provider is now able to verify unearned volumes with this measurement.

\noindent\textbf{How to get the unlearning measurement?}
Given a converged ML model, model sensitivity of one class is with upper and lower bounds. Within this interval, the model sensitivity decreases approximately linearly with increasing sample size (validations are given in Section \ref{subsubsec:exp-Rationale-Metric-II}). 
Generally, we use shadow models to obtain the unlearning measurement, which process is depicted in Figure~\ref{Fig5:Unlearning-Metric-II}. We now elaborate on this process.

The data provider trains $n$ shadow models based on target data. Shadow model training can cost additional computation overhead in Unlearning-Metric-II. We note that this is tolerable once privacy is a top priority for the data provider. Before training, the target class of the unlearned data is divided into different slices according to the volume of data. Taking the CIFAR-10 dataset as an example, when the data contributor's target class is the ``dog'', the volume of the ``dog'' of the requested data is divided incrementally as a shadow dataset, such as $D_{\rm{shadow_1}}^{\rm{dog}}$= 100, $D_{\rm{shadow_2}}^{\rm{dog}}$= 200, ..., $D_{\rm{shadow_n}}^{\rm{dog}}$= 1000. The rest of the categories remain the same amount of data in all the shadow models and do not need to have any intersection with the training set categories in the origin model $M_o$. The data provider then extracts the model sensitivity of class ``dog'' in these shadow models and computes the model sensitivity difference:
\begin{equation}
DS'_{n}=MS_{\rm{shadow_{n+1}}}- MS_{\rm{shadow_n}}. 
\end{equation}
In Metric-II, we define average value of model sensitivity differences in shadow models as unlearning measurement: 
\begin{equation}
UM_{\rm{batch}}=\frac{\sum_{i=1}^{n}DS'_{\rm{2i-1}}}{n}.
\end{equation}

To this end, the data provider is able to use $UM_{\rm{batch}}$ to verify the requested data volume forgotten by the server according to:
\begin{equation}
    Target~data~volume=\left \lceil \frac{DS}{UM_{\rm{batch}} }  \right \rceil \times batch~volume.
\end{equation}

\textcolor{blue}\noindent\textbf{Rationale of Unlearning-Metric-II.} We extract the model sensitivity changes of different unlearned volumes. We found that the model sensitivity changes approximately linearly with the unlearned data volume, which is quantified to compute the Unlearning Measurement. We utilize it to infer the volume of unlearned data in Unlearned Volume Verification. Details are deferred to Section \ref{subsubsec:exp-Rationale-Metric-II}.

\subsection{Unlearning-Metric-III: Unlearned Sample Verification} \label{subsec:Unlearning-Metric-III} 
\begin{figure}[h]
  \centering
  \includegraphics[width = 0.45\textwidth]{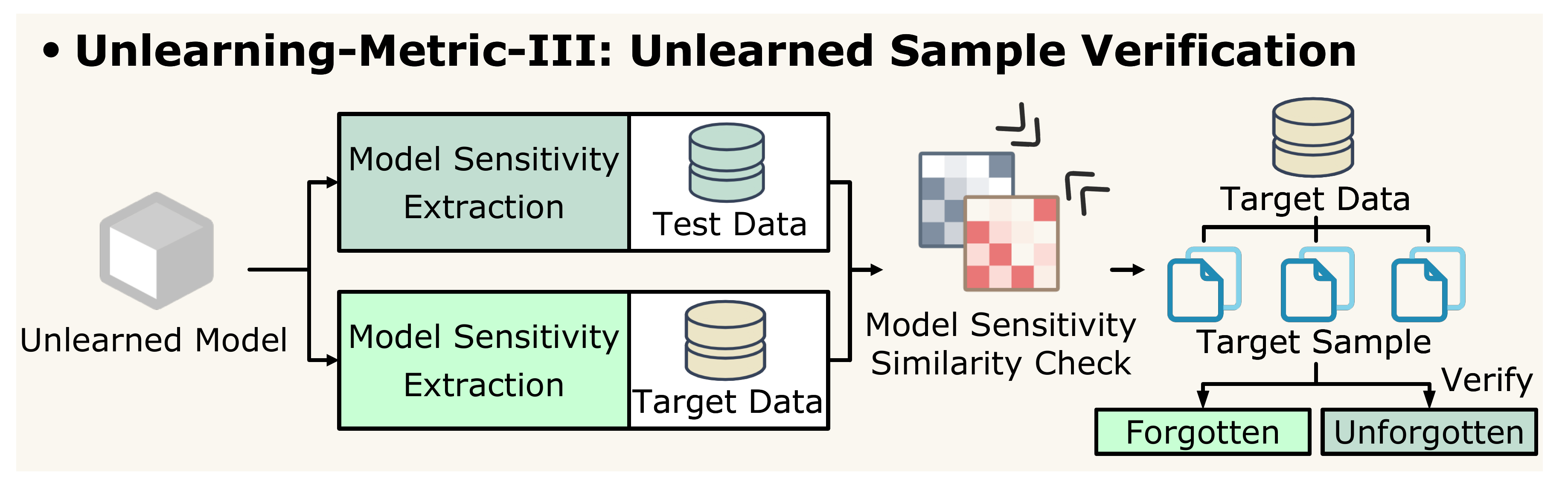}
  \caption{Process of Unlearned Sample Verification.}
  \label{Fig6:Unlearning-Metric-III}
\end{figure}

Unlearning-Metric-I and Unlearning-Metric-II allow data provider to verify the data facticity and volume integrity of machine unlearning in terms of target class and volume. However, in the case of the Deceiving Server, the unlearned class and volume can be the same as the target data, which obscures Unlearning-Metric-I and -II. For example, when data contributor A requests to unlearn 100 samples in the class ``dog'', the Deceiving Server responds through forgetting another non-overlapped 100 samples in the class ``dog'' instead of 100 samples owned by data contributor A. In this case, Unlearning-Metric-I and -II could hardly identify the dishonest behavior of the Deceiving Server, because the server could bypass the verifications on the granularities of class and volume. Hence, we propose Unlearned Sample Verification to deal with such a situation under Sample Granularity, to verify the target sample non-existence in the unlearned model $M_u$.

Unlearned Sample Verification is again based on model sensitivity. The data provider extracts the model sensitivity of the origin model $M_o$ and unlearned model $M_u$ by target data respectively. Through our key observation, model sensitivity of forgotten data is obviously higher than the unforgotten data's model sensitivity when extracted by the target data. Since the model sensitivity measures the familiarity of the model with the samples, when the model trains the unforgotten samples again, it will not have significant gradient changes to the model during back-propagation because the model has already learned their features, and therefore the model sensitivity will not change significantly. 
Conversely, for the forgotten samples, the new features brought to the model will enforce the model to converge toward them, which results in a notable gradient change. This is the foundation of the learning ability of the neural network. Thus, forgotten samples have a more significant change in model sensitivity than that of non-forgotten samples. We leverage this observation to design the Unlearning-Metric-III, and its process is shown in Figure~\ref{Fig6:Unlearning-Metric-III}.

Specifically, the data provider extracts the model sensitivity of unlearned model $MS_{\rm{u\_test}}$ based on test data and $MS_{\rm{u\_tar}}$ based on target data. By comparing $MS_{\rm{u\_test}}$ and $MS_{\rm{u\_tar}}$, if there is an obvious discrepancy between them, then there is a high probability that the server has spoofed the data provider and data contributor (detailed in Fig \ref{subsubsec:exp-Rationale-Metric-III}).  Otherwise, the server has unlearned the target data honestly when $MS_{\rm{u\_test}}$ and $MS_{\rm{u\_tar}}$ are similar. The reason is that the existence of target sample(s) in unlearned model $M_u$ would mitigate gradient changes during the model sensitivity extraction under target data. It could cause a large gap in the model sensitivity by the test data. 
Hence, when the Deceiving Server unlearns the same volume of \textit{irrelevant} data as the target data, it leaves traceable model sensitivity. Detailed analyses are in Section \ref{subsubsec:exp-Rationale-Metric-III}. 

\noindent\textbf{Necessity of target data and test data.}
For honest unlearning, both the target data and test data effect on gradient change will be removed in the unlearned model. Equally, they are unseen data to honestly unlearned model. Therefore, using same amount of target data and test data to extract the model sensitivity of the unlearned model respectively will result in similar gradient change tendency. In contrast, the Deceiving Server aims at reserving all or partial target data in the unlearned model by replacing same amount of non-overlapped data to the requested data to forget. When performing the above processing on the unlearned model, Deceiving Server reserved (partial) target data could mitigate the gradient changes in the model sensitivity extracted with all the target data (in particular, the requested data), rendering lower model sensitivity (detailed in Figure \ref{Fig13:UM-III}). So that quantifying the absolute sensitivity change becomes obscure. To amplify the sensitivity gap, we utilize the sensitivity change extracted upon the test data as a reference to form a differential sensitivity change compared to the sensitivity extracted upon the target data. Test data serves as a good reference because it is unseen by neither the honest server nor the Deceiving Server.

\noindent\textbf{Rationale of Unlearning-Metric-III.} We extract the model sensitivity of the Honest Server and the Deceiving Server on the target data and testing data respectively. We found that, the model sensitivity extracted on target data is significantly lower than the one extracted on testing data when target data remains in the unlearned model. This is the foundation of determining whether the target data still exists within the unlearned model in Unlearned Sample Verification. Details are deferred to Section \ref{subsubsec:exp-Rationale-Metric-III}.

Note that, Unlearning Sample Verification is distinct from the membership inference. Although membership inference could indirectly determine the data record existence, it is not delicately suitable for unlearning verification. Moreover, its accuracy is lower than \name, since membership inference is heavily overfitting-dependent, and we have evaluated and compared with it in Section \ref{subsec:UMIII-mem}.

With the above three \name Unlearning-Metrics or took-kits we propose, the data provider can effectively verify the data facticity and volume integrity of machine unlearning to identify the three types of dishonest servers (Neglecting Server, Lazy Server, and Deceiving Server), and safeguard the privacy and property of machine unlearning users.

\section{Experiments} \label{sec:experiments}
In this Section, we experimentally evaluate \name to validate its effectiveness. We also conduct empirical experiments of \name to affirm its immediate applicability under state-of-the-art machine unlearning frameworks. 
\subsection{Experimental Setup} \label{subsec:experimental setup} 

\subsubsection{Dataset} We run experiments on three common benchmark datasets that have also been used in machine unlearning studies:

\begin{itemize}[leftmargin=*]
    \item \textbf{CIFAR-10} is an image dataset for the recognition of universal objects~\cite{CIFAR10}. There are 10 categories of RGB color images	with size of $32 \times 32 \times 3$. It consists of 50,000 training and 10,000 testing samples, respectively.
    \item \textbf{Fashion-MNIST} is a collection of Zalando's fashion objects, having a training set of 60,000 examples and a test set of 10,000 examples \cite{Fashion-MNIST}. Each sample is a $28  \times 28$ grayscale image, associated with a label from 10 classes.
    \item \textbf{RAF-DB} is a large-scale facial expression database with around 30K great-diverse real-world facial images, called Real-world Affective Faces Database~\cite{RAF-DB}. 
    There are 7 categories of RGB color images with size of $48 \times 48$. 
\end{itemize}

\subsubsection{Model Structure}

As for the CIFAR-10, the model architecture has four convolution blocks: each block has one convolutional layer and one max pooling layer, followed by one fully connected layer. As for the Fashion-MNIST, the model architecture is two convolutional layers and one max pooling layer, followed by two fully connected layers. As for the RAF-DB, the model architecture has four convolution blocks identical to CIFAR-10, with the last layer followed by two fully connected layers.

\subsubsection{Machine Unlearning Method}
In our experiments, we first used machine unlearning with retraining from scratch as the validation baseline of our proposed metrics. Because both exact unlearning and approximate unlearning are extended based on retraining \cite{ThudiJSP22}. In addition, to demonstrate the generalisability of our verification approach, we further evaluate \name under existing SOTA machine unlearning frameworks: the exact unlearning framework SISA \cite{Bourtoule2021} and the approximate unlearning framework Amnesiac Machine Learning \cite{GravesNG21}, respectively.

\subsubsection{Machine Configuration} Experiments run on a computer with the following configuration: Intel Core i9 processor with ten CPU cores running at 3.70 GHz and with a 32 GB main memory, and a GPU card of NVIDIA GeForce RTX 3090.

\subsection{Verification on Unlearning-Metric-I} \label{subsec:exp-Unlearning-Metric-I}
Here, we carry out experiments to evaluate the performance of the Unlearning-Metric-I against the Neglecting Server. The Neglecting Server would ignore the data provider's unlearning request. 
We compare the model sensitivity of target class between the origin model $M_o$ and the unlearned model $M_u$. If the difference of these twin models is lower than a threshold of, e.g., 1$\%$. 
\name judges that the model sensitivity remains unchanged, which means that the server does not honestly unlearn this target class.

\begin{table}[htb]
\centering
\caption{Verification Efficiency of Unlearning-Metric-I}
\label{tab2:UM-I-effiency}
\resizebox{0.60\columnwidth}{!}{
\begin{tabular}{c|cc|c}
\hline
\rowcolor[HTML]{FFFFFF} 
\textbf{Task}                                           & \textbf{Trial}              & \textbf{Vef. Acc.} & \textbf{Overall}                               \\ \hline
\rowcolor[HTML]{FFFFFF} 
\cellcolor[HTML]{FFFFFF}                                & \cellcolor[HTML]{E5E5E5}20  & \cellcolor[HTML]{E5E5E5}0.94   & \cellcolor[HTML]{FFFFFF}                       \\
\rowcolor[HTML]{FFFFFF} 
\cellcolor[HTML]{FFFFFF}                                & 50                          & 0.92                           & \cellcolor[HTML]{FFFFFF}                       \\
\rowcolor[HTML]{FFFFFF} 
\cellcolor[HTML]{FFFFFF}                                & \cellcolor[HTML]{E5E5E5}100 & \cellcolor[HTML]{E5E5E5}0.93   & \cellcolor[HTML]{FFFFFF}                       \\
\rowcolor[HTML]{FFFFFF} 
\multirow{-4}{*}{\cellcolor[HTML]{FFFFFF}CIFAR-10}      & 200                         & 0.92                           & \multirow{-4}{*}{\cellcolor[HTML]{FFFFFF}0.93} \\ \hline
\rowcolor[HTML]{FFFFFF} 
\cellcolor[HTML]{FFFFFF}                                & \cellcolor[HTML]{E5E5E5}20  & \cellcolor[HTML]{E5E5E5}0.95   & \cellcolor[HTML]{FFFFFF}                       \\
\rowcolor[HTML]{FFFFFF} 
\cellcolor[HTML]{FFFFFF}                                & \cellcolor[HTML]{FFFFFF}50  & 0.93                           & \cellcolor[HTML]{FFFFFF}                       \\
\rowcolor[HTML]{FFFFFF} 
\cellcolor[HTML]{FFFFFF}                                & \cellcolor[HTML]{E5E5E5}100 & \cellcolor[HTML]{E5E5E5}0.93   & \cellcolor[HTML]{FFFFFF}                       \\
\rowcolor[HTML]{FFFFFF} 
\multirow{-4}{*}{\cellcolor[HTML]{FFFFFF}Fashion-MNIST} & \cellcolor[HTML]{FFFFFF}200 & 0.92                           & \multirow{-4}{*}{\cellcolor[HTML]{FFFFFF}0.93} \\ \hline
\rowcolor[HTML]{FFFFFF} 
\cellcolor[HTML]{FFFFFF}                                & \cellcolor[HTML]{E5E5E5}20  & \cellcolor[HTML]{E5E5E5}0.90     & \cellcolor[HTML]{FFFFFF}                       \\
\rowcolor[HTML]{FFFFFF} 
\cellcolor[HTML]{FFFFFF}                                & \cellcolor[HTML]{FFFFFF}50  & 0.88                             & \cellcolor[HTML]{FFFFFF}                       \\
\rowcolor[HTML]{FFFFFF} 
\cellcolor[HTML]{FFFFFF}                                & \cellcolor[HTML]{E5E5E5}100 & \cellcolor[HTML]{E5E5E5}0.91     & \cellcolor[HTML]{FFFFFF}                       \\
\rowcolor[HTML]{FFFFFF} 
\multirow{-4}{*}{\cellcolor[HTML]{FFFFFF}RAF-DB}        & \cellcolor[HTML]{FFFFFF}200 & 0.90                             & \multirow{-4}{*}{\cellcolor[HTML]{FFFFFF}0.89}   \\ \hline
\end{tabular}}
\end{table}

\noindent \textbf{Metric-I Verification Accuracy.} We first conduct evaluations on baseline honest unlearning and dishonest unlearning by retraining from scratch to verify the unlearning effectiveness. We have repeatedly run 20, 50, 100 and 200 experiments to determine whether the target class is unlearned in unlearned model $M_u$, and reported verification accuracy of Unlearning-Metric-I as shown in Table \ref{tab2:UM-I-effiency}. The experimental results indicate that Unlearning-Metric-I can reliably judge the data class forgotten by the server with high accuracy, so it has the ability to verify the Neglecting Server.

\begin{figure}[htb]
    \centering
	\subfigure[Unlearned Class Number]{
		\includegraphics[width = 0.21\textwidth]{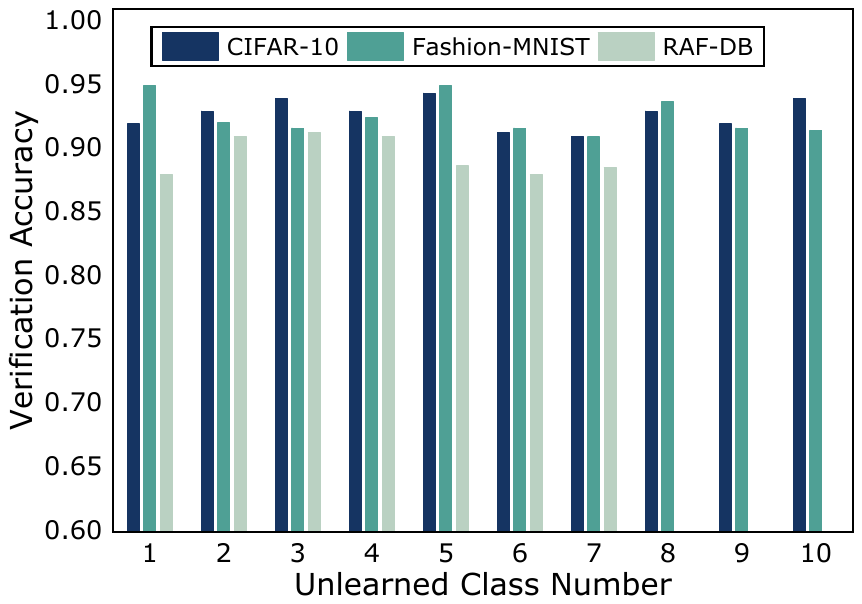}
		
	}
	\subfigure[Unlearned Sample Volume]{
		\includegraphics[width = 0.21\textwidth]{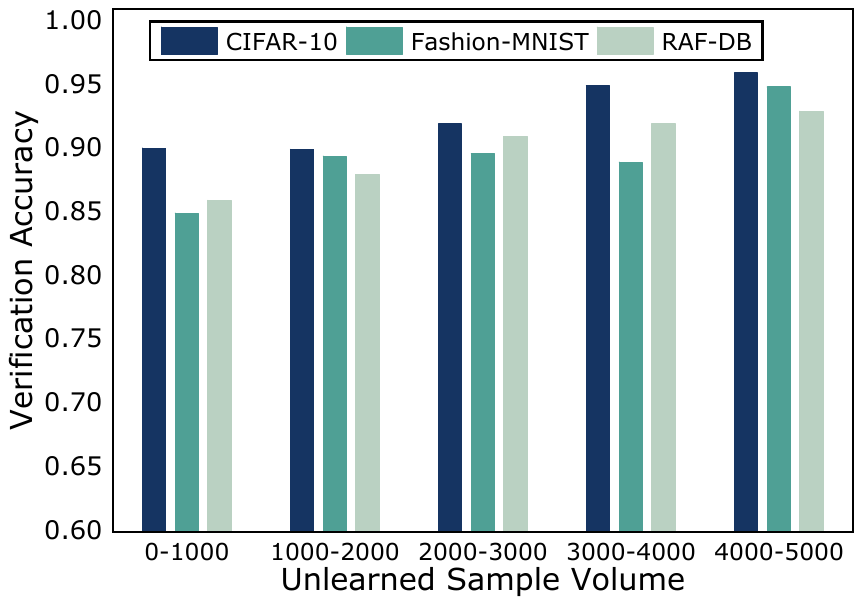}
	}
	\caption{Unlearning-Metric-I's performance under different class numbers and sample volumes.} 
	\label{Fig7:UM-1-class-sample}
\end{figure}

\noindent \textbf{Impact of Class Number and Sample Volume.} 
Here, we show the verification ability of Unlearning-Metric-I towards different unlearning requests. We select 1-10 unlearned class(es) and 0-5000 unlearned samples given a class (1-7 classes for RAF-DB, since it has 7 classes) which represents the variety of unlearning requests. Under these settings, we test the accuracy of Unlearning-Metric-I, and the results are shown in Figure \ref{Fig7:UM-1-class-sample}.

As can be seen from Figure \ref{Fig7:UM-1-class-sample} (a) and (b), verification accuracy of Unlearning-Metric-I can maintain more than 90\% for CIFAR-10 and Fashion-MNIST, 85\% for RAF-DB  under different unlearned classes and sample volumes. In terms of unlearned classes, Unlearning-Metric-I accuracy is insensitive to it. In terms of sample size, the accuracy of the verification increases slightly as the number of unlearned samples increases, which is because the change in model sensitivity is related to the amount of data within the unlearned model $M_u$, i.e. the remaining amount of data in a given class is small. When the amount of data in the target class is smaller, the more obvious gradient changes are produced when extracting model sensitivities.

\subsection{Verification on Unlearning-Metric-II} \label{subsec:exp-Unlearning-Metric-II}
Here, we explore unlearning verification in the Volume Granularity, and infer the amount of target data that the server forgets. \name can be used to counteract the behavior of Lazy Server that unlearns only partial target data.

\begin{table}[htb]
\centering
\caption{Verification Deviation of Unlearning-Metric-II}
\label{tab3:UM-II-deviation}
\resizebox{0.75\columnwidth}{!}{
\begin{tabular}{c|ccc|c}
\hline
\rowcolor[HTML]{FFFFFF} 
\textbf{Task}                                                                                      & \textbf{\begin{tabular}[c]{@{}c@{}}Unlearned \\ Volume\end{tabular}} & \textbf{\begin{tabular}[c]{@{}c@{}}Inferred \\ Volume\end{tabular}} & \textbf{Deviation} & \textbf{Overall}                                \\ \hline
\rowcolor[HTML]{E5E5E5} 
\cellcolor[HTML]{FFFFFF}                                                                           & 500                                                                  & 470                                                                 & 6\%                & \cellcolor[HTML]{FFFFFF}                        \\
\rowcolor[HTML]{FFFFFF} 
\cellcolor[HTML]{FFFFFF}                                                                           & 1000                                                                 & 942                                                                 & 5.8\%              & \cellcolor[HTML]{FFFFFF}                        \\
\rowcolor[HTML]{E5E5E5} 
\cellcolor[HTML]{FFFFFF}                                                                           & 1500                                                                 & 1421                                                                & 5.3\%              & \cellcolor[HTML]{FFFFFF}                        \\
\rowcolor[HTML]{FFFFFF} 
\multirow{-4}{*}{\cellcolor[HTML]{FFFFFF}CIFAR-10}                                                 & 2500                                                                 & 2443                                                                & 2.3\%              & \multirow{-4}{*}{\cellcolor[HTML]{FFFFFF}4.8\%} \\ \hline
\rowcolor[HTML]{E5E5E5} 
\cellcolor[HTML]{FFFFFF}                                                                           & 500                                                                  & 463                                                                 & 7.4\%              & \cellcolor[HTML]{FFFFFF}                        \\
\rowcolor[HTML]{FFFFFF} 
\cellcolor[HTML]{FFFFFF}                                                                           & 1000                                                                 & 935                                                                 & 6.5\%              & \cellcolor[HTML]{FFFFFF}                        \\
\rowcolor[HTML]{E5E5E5} 
\cellcolor[HTML]{FFFFFF}                                                                           & 1500                                                                 & 1419                                                                & 5.4\%              & \cellcolor[HTML]{FFFFFF}                        \\
\rowcolor[HTML]{FFFFFF} 
\multirow{-4}{*}{\cellcolor[HTML]{FFFFFF}\begin{tabular}[c]{@{}c@{}}Fashion\\ -MNIST\end{tabular}} & 2500                                                                 & 2388                                                                & 4.5\%              & \multirow{-4}{*}{\cellcolor[HTML]{FFFFFF}6.8\%} \\ \hline
\rowcolor[HTML]{E5E5E5} 
\cellcolor[HTML]{FFFFFF}                                                                           & 500                                                                  & 452                                                                 & 9.7\%              & \cellcolor[HTML]{FFFFFF}                        \\
\rowcolor[HTML]{FFFFFF} 
\cellcolor[HTML]{FFFFFF}                                                                           & 1000                                                                 & 912                                                                 & 8.8\%              & \cellcolor[HTML]{FFFFFF}                        \\
\rowcolor[HTML]{E5E5E5} 
\cellcolor[HTML]{FFFFFF}                                                                           & 1500                                                                 & 1388                                                                & 7.5\%              & \cellcolor[HTML]{FFFFFF}                        \\
\rowcolor[HTML]{FFFFFF} 
\multirow{-4}{*}{\cellcolor[HTML]{FFFFFF}RAF-DB}                                                   & 2500                                                                 & 2330                                                                & 6.8\%              & \multirow{-4}{*}{\cellcolor[HTML]{FFFFFF}8.2\%} \\ \hline
\end{tabular}}
\end{table}

\noindent \textbf{Inference of Unlearned Sample Volume.} We first conduct experiments on inferring the number of unlearned samples under the baseline machine unlearning and summarize results in Table \ref{tab3:UM-II-deviation}. We evaluate Unlearning-Metric-II under the scenarios that the unlearned sample volume is 500, 1000, 1500, and 2500 respectively for CIFAR-10 and Fashion-MNIST, and compute that:
\begin{align}
    Deviation=\frac{\left |Unlearned~Volume-Verified~Volume \right | }{Unlearned~Volume}
\end{align}
, which represents the difference degree between the factual requested unlearned volume and the estimated verified volume. The closer the deviation is to 0, the better the verification effect of Unlearning-Metric-II is. We have run each task for 50 trials. From Table \ref{tab3:UM-II-deviation}, the data volume extrapolated from Unlearning-Metric-II is highly reliable, differing from the factual unlearned volume by only 4.8\% for CIFAR-10, 6.8\% for Fashion-MNIST and 8.2\% for RAF-DB on average. Additionally, the verification effect improves with the increase of unlearned volume, due to the more pronounced changes in model sensitivity caused by large data volume changes.

\begin{figure}[htb]
    \centering
	\subfigure[Unlearned Class Number]{
		\includegraphics[width = 0.21\textwidth]{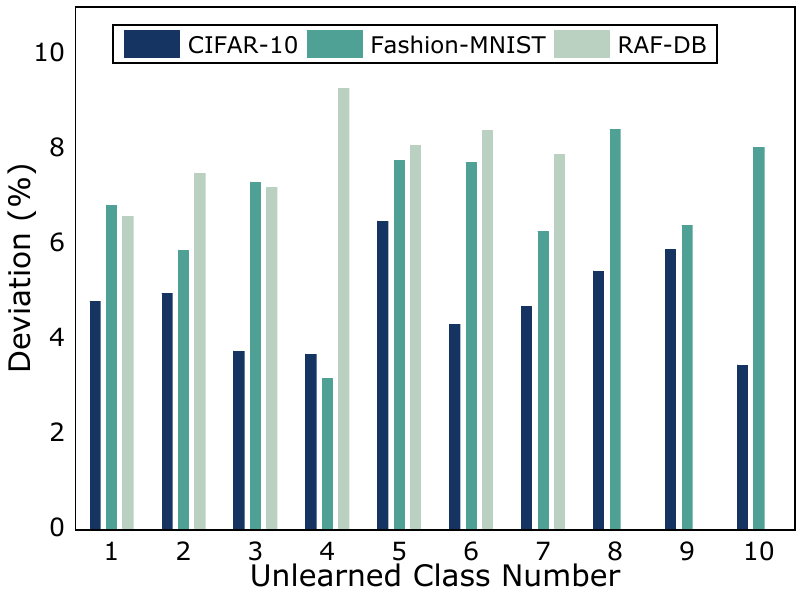}
		
	}
	\subfigure[Shadow Model Number]{
		\includegraphics[width = 0.21\textwidth]{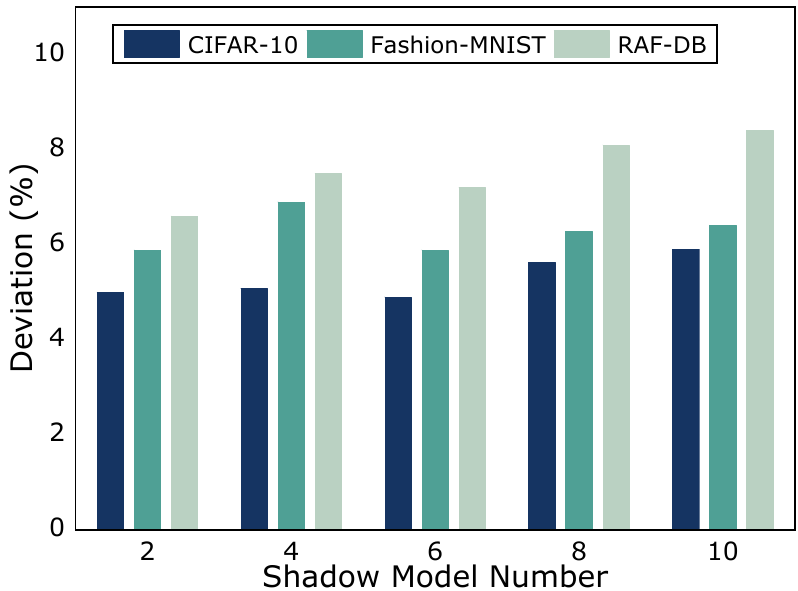}
	}
	\caption{Unlearning-Metric-II's deviation under different unlearned class numbers and shadow models. }
	\label{Fig9: UM-II-class-shadow}
\end{figure}

\noindent \textbf{Impact of Class Number and Shadow Model Number.} We evaluate the Verification Deviation under different unlearned class numbers and shadow model numbers. We set 1-10 (1-7 for RAF-DB) unlearned class(es) (in increments of 1) and 2-10 (2-7 for RAF-DB) shadow models (in increments of 2) in the experiments respectively. With this setting, we conduct multiple experiments and report the deviation of each case in Figure \ref{Fig9: UM-II-class-shadow}. The experimental results demonstrate that the efficiency of Unlearning-Metric-II is invariant to the number of classes and shadow models. The deviation range is small, falling between 3\%-9\% with diverse settings. 
 
\begin{table}[htb]
\centering
\caption{Verification Efficiency of Unlearning-Metric-III}
\label{tab4:UM-III-effiency}
\resizebox{0.63\columnwidth}{!}{
\begin{tabular}{c|cc|c}
\hline
\rowcolor[HTML]{FFFFFF} 
\textbf{Task}                                           & \textbf{Trial}              & \textbf{Vef. Acc.} & \textbf{Overall}                               \\ \hline
\rowcolor[HTML]{FFFFFF} 
\cellcolor[HTML]{FFFFFF}                                & \cellcolor[HTML]{E5E5E5}20  & \cellcolor[HTML]{E5E5E5}0.93   & \cellcolor[HTML]{FFFFFF}                       \\
\rowcolor[HTML]{FFFFFF} 
\cellcolor[HTML]{FFFFFF}                                & 50                          & 0.89                           & \cellcolor[HTML]{FFFFFF}                       \\
\rowcolor[HTML]{FFFFFF} 
\cellcolor[HTML]{FFFFFF}                                & \cellcolor[HTML]{E5E5E5}100 & \cellcolor[HTML]{E5E5E5}0.87   & \cellcolor[HTML]{FFFFFF}                       \\
\rowcolor[HTML]{FFFFFF} 
\multirow{-4}{*}{\cellcolor[HTML]{FFFFFF}CIFAR-10}      & 200                         & 0.90                           & \multirow{-4}{*}{\cellcolor[HTML]{FFFFFF}0.90} \\ \hline
\rowcolor[HTML]{FFFFFF} 
\cellcolor[HTML]{FFFFFF}                                & \cellcolor[HTML]{E5E5E5}20  & \cellcolor[HTML]{E5E5E5}0.94   & \cellcolor[HTML]{FFFFFF}                       \\
\rowcolor[HTML]{FFFFFF} 
\cellcolor[HTML]{FFFFFF}                                & \cellcolor[HTML]{FFFFFF}50  & 0.91                           & \cellcolor[HTML]{FFFFFF}                       \\
\rowcolor[HTML]{FFFFFF} 
\cellcolor[HTML]{FFFFFF}                                & \cellcolor[HTML]{E5E5E5}100 & \cellcolor[HTML]{E5E5E5}0.91   & \cellcolor[HTML]{FFFFFF}                       \\
\rowcolor[HTML]{FFFFFF} 
\multirow{-4}{*}{\cellcolor[HTML]{FFFFFF}Fashion-MNIST} & \cellcolor[HTML]{FFFFFF}200 & 0.92                           & \multirow{-4}{*}{\cellcolor[HTML]{FFFFFF}0.92} \\ \hline
\rowcolor[HTML]{FFFFFF} 
\cellcolor[HTML]{FFFFFF}                                & \cellcolor[HTML]{E5E5E5}20  & \cellcolor[HTML]{E5E5E5}0.90     & \cellcolor[HTML]{FFFFFF}                       \\
\rowcolor[HTML]{FFFFFF} 
\cellcolor[HTML]{FFFFFF}                                & \cellcolor[HTML]{FFFFFF}50  & 0.93                             & \cellcolor[HTML]{FFFFFF}                       \\
\rowcolor[HTML]{FFFFFF} 
\cellcolor[HTML]{FFFFFF}                                & \cellcolor[HTML]{E5E5E5}100 & \cellcolor[HTML]{E5E5E5}0.92     & \cellcolor[HTML]{FFFFFF}                       \\
\rowcolor[HTML]{FFFFFF} 
\multirow{-4}{*}{\cellcolor[HTML]{FFFFFF}RAF-DB}        & \cellcolor[HTML]{FFFFFF}200 & 0.91                             & \multirow{-4}{*}{\cellcolor[HTML]{FFFFFF}0.91}   \\ \hline
\end{tabular}}
\end{table}

\subsection{Verification on Unlearning-Metric-III} \label{subsec:exp-Unlearning-Metric-III}
Here, we evaluated the Unlearning-Metric-III performance, specifically the existence of target sample(s) in the unlearned model $M_u$. This can be utilized to identify the behavior of Deceiving Server that secretly swaps the target sample(s).

\noindent \textbf{Metric-III Verification Accuracy.} 
The purpose of Unlearning-Metric-III is to verify whether the samples that the server has unlearned are the target samples in the unlearning request. Then we evaluate the verification efficiency of Unlearning-Metric-III under the scenario with honest unlearning and dishonest unlearning (Deceiving Server). We conducted the experiments for 20, 50, 100 and 200 different trial runs. Table \ref{tab4:UM-III-effiency} illustrates the verification accuracy and overall accuracy. Unlearning-Metric-III could maintain an accuracy of 90\% on average, which provides the data contributor with evidence to verify the existence of his/her target data.

\begin{figure}[htb]
	\centering
	\subfigure[Unlearned Sample Volume]{
		\includegraphics[width = 0.205\textwidth]{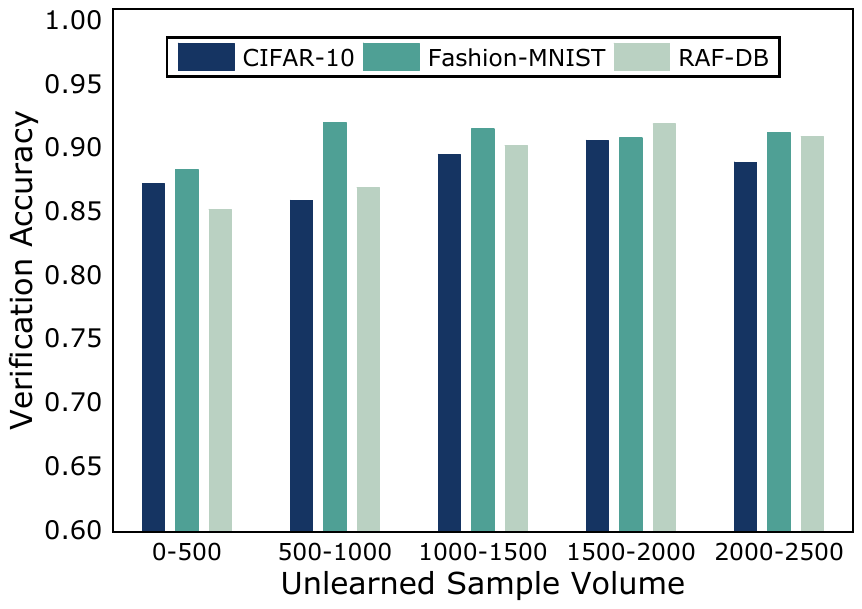}
		
	}
	\subfigure[Unlearned Class Number]{
		\includegraphics[width = 0.215\textwidth]{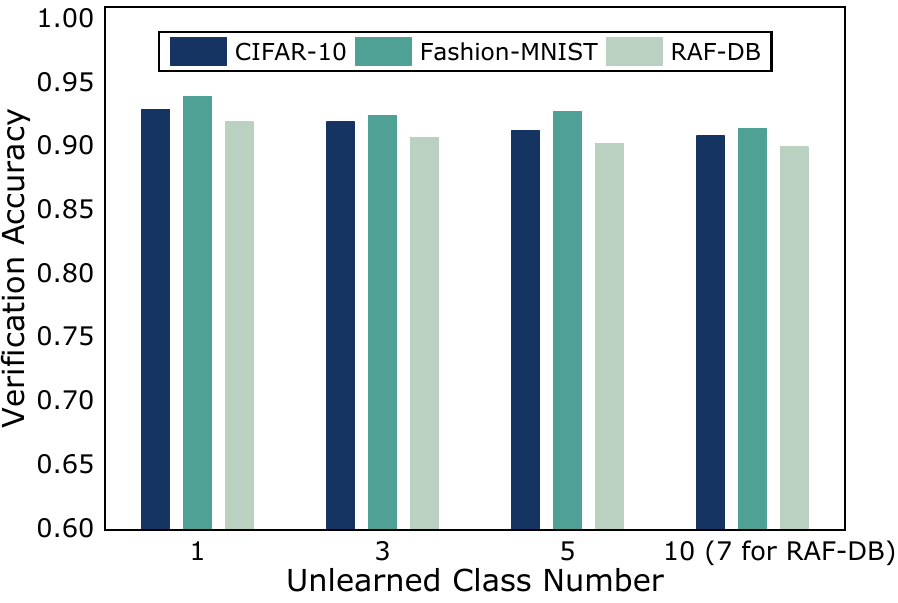}
	}
	\caption{Unlearning-Metric-III's performance under different sample volumes and class numbers. }
	\label{Fig10:UM-III-sample}
\end{figure}

\noindent \textbf{Impact of Sample Volume and Class Number.} 
Figure \ref{Fig10:UM-III-sample} (a) depicts the verification accuracy under different amounts of target data, which is from 0-2500 for CIFAR-10 and Fashion-MNIST (the server needs the rest 2500 to replace the target data in unlearning request). The experimental results depict that the verification efficiency is minimally affected by the number of unlearned samples. Meanwhile, we also detail the verification accuracy when the class number is changed in Figure \ref{Fig10:UM-III-sample} (b). We select 1, 3, 5, 10 (7 for RAF-DB dataset, because it has 7 classes) unlearned classes respectively, and the result of Unlearning-Metric-III can maintain more than 90\% verification accuracy to identify the Deceiving Server.

\subsection{Unlearning Frameworks Verification}  \label{subsec:exp-Unlearning Verification Generalisability} 
The machine unlearning method we have evaluated so far is the baseline machine unlearning framework that trains the entire model from scratch. Here we evaluate \name on existing SOTA unlearning frameworks specifically, SISA \cite{Bourtoule2021} as the representative of exact unlearning, and Amnesiac Unlearning \cite{GravesNG21} as the representative of approximate unlearning. We aim to validate the immediate applicability of \name for available or potentially emerging unlearning frameworks. For each unlearning framework, we repeated 10 trials to evaluate three unlearning metrics.

\noindent\textbf{Exact Unlearning Setting.}
The core of SISA~\cite{Bourtoule2021} is to split the training data into $k$ disjoint shards. The server then trains $k$ sub-models respectively and derives the final output by collaborative predictions of all sub-models through aggregating algorithms---an ensemble strategy. When receiving the unlearning request, the server localizes the shards where the target data exist in, and only retrain sub-models corresponding to these shards. 
Such an approach substantially reduces computational overhead than baseline retraining, thus accelerating the unlearning process. In the experiments, we set $k$ = 5 and assign 500 data samples to each shard. We randomly select 0-400 data samples in a shard to unlearn for each trial and evaluate the unlearning accuracy using the proposed Unlearning-Metric-I,-II, and -III. 

\noindent\textbf{Approximate Unlearning Setting.} Meanwhile, we also evaluate \name under approximate unlearning. The goal of approximate unlearning is to muddy the understanding of target data in the model by relabeling the target data with randomly selected incorrect labels and then continue training the model on the modified dataset \cite{GravesNG21}. This behavior could reduce the generalization of the target data. In this experiment, we randomly replace the labels of the target data with others, and continue updating the model on the relabeling dataset. The data volume in origin model $M_o$ is 1500 per class, and the data volume in unlearning request is 0-1500 for a randomly selected class. 

\noindent\textbf{Results.} Table \ref{tab6:UnlearningFramework} details the verification performance (Unlearning-Metric-I, -II, -III) under exact and approximate unlearning, respectively, on the CIFAR-10, Fashion-MNIST and RAF-DB. The experimental results of Unlearning-Metric-I under both unlearning frameworks validate that the verification performance can derive a similar effect to the baseline retraining, i.e.,  the Unlearned Class Verification. Both unlearning frameworks can achieve more than 90\% verification accuracy. The Unlearned Volume Verification and the Unlearned Sample Verification slightly degrade compared to the baseline unlearning, relatively notable for the approximate unlearning. We analyze the reason is that these unlearning frameworks may sacrifice a unlearned model's accuracy for the sake of reducing computation overhead, e.g., when the number of shards increases, it induces accuracy degradation \cite{Bourtoule2021}. It could suppress the normal gradient changes as it could be in the baseline unlearning. Recall that the approximate unlearning directly modifies the parameters of the trained model, which attempts to approximate what the model would have looked like if the data points to be unlearned were not in the training dataset from the beginning \cite{ThudiJSP22}. Such approaches lead to smaller data volume changes. Nevertheless, the mitigation of data influence (mainly the relabeling in Amnesiac Unlearning \cite{GravesNG21}) exactly improves the model sensitivity of unlearned classes, which can be fed in the Unlearning-Metric-II to infer the volume of muddy data partially. Compared with baseline and exact unlearning \cite{Bourtoule2021}, there would exist a certain backward of verification deviation in approximate unlearning. Overall, although the performance of \name on existing SOTA unlearning frameworks sees a slight drop attributing to the potential accuracy trade-off of these unlearning, it remains to be satisfactory.

\begin{table*}[htb]
\centering
\caption{Verification performance of SISA and Amnesiac Unlearning Frameworks}
\label{tab6:UnlearningFramework}
\resizebox{1.70\columnwidth}{!}{
\begin{tabular}{c|c|c|c|c}
\hline
\textbf{Unlearning Framework}                                                                            & \textbf{Task}                      & \textbf{\begin{tabular}[c]{@{}c@{}}Unlearning-Metric-I\\ (Verification Accuracy)\end{tabular}} & \textbf{\begin{tabular}[c]{@{}c@{}}Unlearning-Metric-II\\ (Verification Deviation)\end{tabular}} & \textbf{\begin{tabular}[c]{@{}c@{}}Unlearning-Metric-III\\ (Verification Accuracy)\end{tabular}} \\ \hline
& \cellcolor[HTML]{E5E5E5}CIFAR-10      & \cellcolor[HTML]{E5E5E5}0.92                                                                   & \cellcolor[HTML]{E5E5E5}6.1\%                                                                    & \cellcolor[HTML]{E5E5E5}0.85                                                                     \\
& Fashion-MNIST                         & 0.90                                                                                     & 9.1\%                                                                                     & 0.84                                                                                     \\
\multirow{-3}{*}{\begin{tabular}[c]{@{}c@{}}SISA\\ (Exact Unlearning)\end{tabular}}                      & \cellcolor[HTML]{E5E5E5}RAF-DB        & \cellcolor[HTML]{E5E5E5}0.89                                                             & \cellcolor[HTML]{E5E5E5}9.6\%                                                              & \cellcolor[HTML]{E5E5E5}0.86                                                             \\ \hline
& CIFAR-10                              & 0.94   & 10.1\%                                                                                           & 0.90                                                                    \\
   & \cellcolor[HTML]{E5E5E5}Fashion-MNIST & \cellcolor[HTML]{E5E5E5}0.92                                                              & \cellcolor[HTML]{E5E5E5}10.7\%                                                               & \cellcolor[HTML]{E5E5E5}0.88                                                              \\
\multirow{-3}{*}{\begin{tabular}[c]{@{}c@{}}Amnesiac Unlearning\\ (Approximate Unlearning)\end{tabular}} & RAF-DB                                & \cellcolor[HTML]{FFFFFF}0.88                                                              & \cellcolor[HTML]{FFFFFF}12.1\%                                                             & \cellcolor[HTML]{FFFFFF}0.90                                                              \\ \hline
\end{tabular}}
\end{table*}

\section{Further Evaluation and Discussion} \label{sec:discussion}

\subsection{Unlearning-Metric-III vs. Membership Inference Attack} \label{subsec:UMIII-mem}
Unlearning-Metric-III aims to determine the existence of the target data in the unlearned model $M_u$, ensuring that the data ownership forgotten by the server is distinguishable. It could expose the displacement behavior that retains the target data in unlearned model $M_u$ by the dishonest server. 
Membership inference can serve as an indirect method to non-invasively verify the sample-granularity unlearning to some extent. Therefore, we compare \name with it in terms of sample-granularity verification, affirming that \name is more competent in explicitly verifying the unlearning task.

We utilized the membership inference attack against the machine unlearning \cite{ChenCCS2021} to compare with the Unlearning-Metric-III. In this experiment, we set 1 shadow origin model and 20 shadow unlearned models with 30000 samples for each model. Then, we construct the membership feature by Euclidean distance and train an  attack model through the logistic regression, which is aligned with \cite{ChenCCS2021}. We evaluated the verification accuracy and computation overhead under overfitting and non-overfitting, as shown in Table \ref{tab7:MEM}. The results report that although the membership inference could verify the existence of an  unlearned sample in the unlearned model, it is heavily overfitting-dependent---this is actually a notable inherent downside of membership inference~\cite{hu2022membership}. Additionally, the training time for multiple shadow models also hampers verification efficiency. In the experiment, the computation overhead is 8109.92s, which includes training 1 shadow origin model for 509.96s, training 20 shadow unlearned models with an average of 322.75s per model, constructing features for 1142.41s, and training attack models for 2.49s. This overhead of the membership-inference-based method is much longer compared to the model sensitivity extraction time (52.08s) in \name for all requested samples (1500 samples in this trial).

\begin{table}[htb]
\centering
\caption{Comparison between membership inference attack and Unlearning-Metric-III}
\label{tab7:MEM}
\resizebox{0.8\columnwidth}{!}{
\begin{tabular}{c|cc|c}
\hline
                   & \multicolumn{2}{c|}{\textbf{Membership Inference Attack}}                          &                                   \\ \cline{2-3}
\multirow{-2}{*}{} & \multicolumn{1}{c|}{\textbf{Overfitting}}          & \textbf{Non-Overfitting}      & \multirow{-2}{*}{\textbf{\name}} \\ \hline
Train Acc.         & \multicolumn{1}{c|}{\cellcolor[HTML]{E5E5E5}0.954} & \cellcolor[HTML]{E5E5E5}0.942 & \cellcolor[HTML]{E5E5E5}-         \\ \hline
Test Acc.          & \multicolumn{1}{c|}{0.477}                         & 0.919                         & -                                 \\ \hline
Verf. Acc.          & \multicolumn{1}{c|}{\cellcolor[HTML]{E5E5E5}0.881} & \cellcolor[HTML]{E5E5E5}0.519 & \cellcolor[HTML]{E5E5E5}0.92      \\ \hline
Overhead           & \multicolumn{2}{c|}{8109.92s}                                                       & 52.08s                             \\ \hline
\end{tabular}}
\end{table}

\subsection{Overhead}
Here, we evaluated the computational cost of \name for face recognition tasks using the VGG model. We record the computational costs associated with each dominant operation within the three unlearning metrics.

\begin{itemize}[leftmargin=*]
\item For Unlearning-Metric-I, the data provider needs to perform two model sensitivity extraction operations for one unlearning verification task, with an average time consumption of 90.83 seconds.

\item For Unlearning-Metric-II, the computational cost mainly involves shadow model training and sensitivity extraction. We found that the training time for each shadow model is 716.79 seconds, and sensitivity extraction takes 55.47 seconds. It's worth noting that shadow model training is the dominant factor in computational cost, as it involves extracting the correlation between sample size and model sensitivity. However, in real-world applications, shadow model training is a one-time process and doesn't need to be repeated for each unlearning verification. Additionally, the training can be conducted offline before the verification task, reducing user waiting time.

\item For Unlearning-Metric-III, the computational cost relies on two model sensitivity extractions, with an average computational cost of 45.33 seconds per extraction. Specifically, we conducted a detailed analysis and comparison of the performance between Unlearning-Metric-III and membership inference, as shown in Table \ref{tab7:MEM}.

\end{itemize}

Therefore, \name meets the computational cost requirements for real-world applications, as it doesn't impose a significant burden on the data provider and avoids prolonged waiting times for data contributors.

\subsection{Limitation}
\name focuses only on the unlearning verification of classification tasks, primarily in the field of image classification. This is aligned with most of the existing unlearning verification methods~\cite{ShokriSSS17,SongS19,ChenCCS2021,gao2020backdoor}, which are mainly focused on classification tasks. There is limited research on unlearning in nonclassification tasks~\cite{XuZZZY24}, such as regression and generative models (i.e., LLM). These tasks may present higher model complexity and varying modalities, making it more challenging to perform unlearning. At the same time, it poses new challenges to apply the verification of unlearning, including \name, which tends to be an open challenge.

\section{Conclusions} \label{sec:conclusion}
This work is an initiative towards the machine unlearning verification framework entailed with triple degrees of granularity (class-, volume-, sample-level) to verify the data facticity and volume integrity of machine unlearning, namely \name. We identified three types of dishonest servers (Neglecting Server, Lazy Server, Deceiving Server) and proposed three unlearning metrics to counter them. Specifically, Unlearning-Metric-I verifies if the server unlearned the target class, Unlearning-Metric-II estimates the actual data volume unlearned by the server, and Unlearning-Metric-III confirms the presence of the target sample in the unlearned model.
Extensive experiments have affirmed that the \name is effective under various verification scenarios and significantly, is immediately applicable for existing SOTA unlearning frameworks SISA \cite{Bourtoule2021} as a representative of exact unlearning, and Amnesiac Unlearning \cite{GravesNG21} as a representative of approximate unlearning. We have further demonstrated the impact of \name in real-world applications, particularly in face recognition systems.

\bibliographystyle{plain}
\bibliography{TruVRF}

\appendices

\section{The Rationale of \name} \label{subsec:exp-Rationale}
Here, we conduct evaluations on CIFAR-10 to further explain the \name rationale.
\subsection{\textbf{Rationale of Unlearning-Metric-I}} \label{subsubsec:exp-Rationale-Metric-I}

\begin{figure}[htb]
	\centering
	\includegraphics[width = 0.35\textwidth]{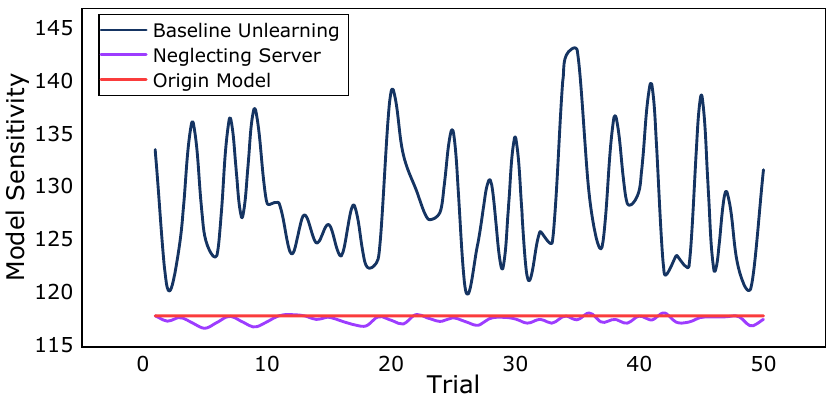}
	\caption{Key observation of Unlearning-Metric-I. The model sensitivity of honest unlearning and dishonest unlearning with Neglecting Server exhibit a significant discrepancy. }
	\label{Fig11:UM-I}
\end{figure}

We elaborate on the key observation supporting Unlearning-Metric-I in Figure \ref{Fig11:UM-I}. In this experiment, we set 5000 samples per class in the origin model $M_o$, and issue 50 unlearning requests, respectively. The target class and the sample number are random. For comparison, we employ a pair of honest unlearning and dishonest unlearning with Neglecting Server per unlearning request, and extract model sensitivity of unlearned model $M_u$. The figure clearly shows that Unlearning-Metric-I can verify whether the server unlearns the target class attributing to the obvious model sensitivity discrepancy, thus distinguishing the honest server from Neglecting Server.

\subsection{\noindent\textbf{Rationale of Unlearning-Metric-II}} \label{subsubsec:exp-Rationale-Metric-II}

\begin{figure}[htb]
	\centering
	\includegraphics[width = 0.35\textwidth]{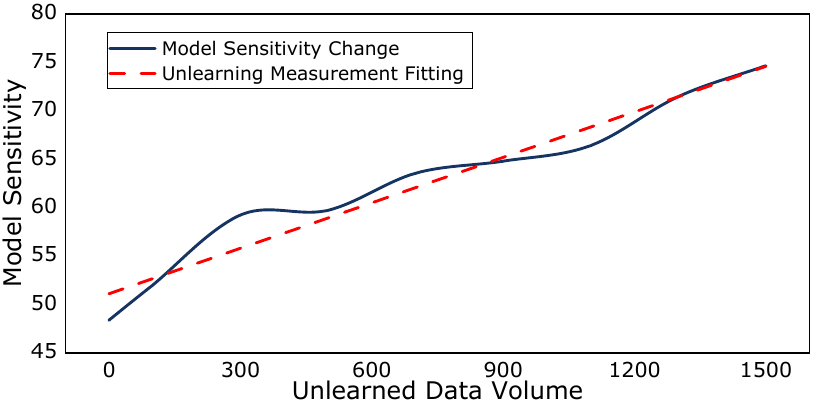}
	\caption{Key observation of Unlearning-Metric-II. The model sensitivity changes approximately linearly with the unlearned data volume representing the variation value of model sensitivity per unit of unlearned data volume, which is denoted as the Unlearning Measurement. We utilize it to infer the scale of unlearned data by the server in Unlearning-Metric-II.}
	\label{Fig12:UM-II}
\end{figure}

To interpret Metric-II and Unlearning Measurement, we depict the model sensitivity changes of target data as unlearning data volume increases in a model in Figure \ref{Fig12:UM-II}. Meanwhile, we fit such changes into a curve to demonstrate the Unlearning Measurement, that is, how much data does a fixed amount of model sensitivity change correspond to. 
In this experiment, we increase unlearning requested samples from 0 to 1500. As Figure \ref{Fig12:UM-II} validates, the increment of unlearning data volume leads to a linear increase of model sensitivity of the corresponding class.

\subsection{\noindent\textbf{Rationale of Unlearning-Metric-III}} \label{subsubsec:exp-Rationale-Metric-III}
\begin{figure}[htb]
	\centering
	\includegraphics[width = 0.35\textwidth]{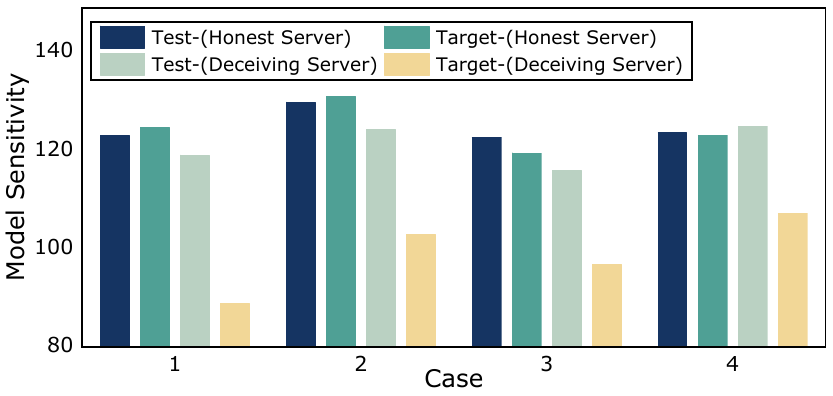}
	\caption{Key observation of Unlearning-Metric-III. The four types of model sensitivities: Test-(Honest Server) and Target-(Honest Server) represent the model sensitivity extracted by test data and target data in Honest Server scenario, and Test-(Deceiving Server) and Target-(Deceiving Server) represent the model sensitivity extracted by test data and target data in Deceiving Server scenario.}
	\label{Fig13:UM-III}
\end{figure}

To interpret the rationale behind Unlearning-Metric-III, we show the difference in model sensitivity under Honest and Deceiving Server. We set the Honest Server and the Deceiving Server as the control group. Following Algorithm \ref{alg4:Unlearning-Metric-III}, we extract the model sensitivity of unlearned model $M_u$ using test data and target data. In an experimental setting, the size of the data contributor target class is 5000, the unlearning rate is randomly selected from 4\% to 20\%, and we perform four cases. Figure \ref{Fig13:UM-III} shows the comparison of the model sensitivity. For the same unlearned model $M_u$, the model sensitivity extracted by test data (i.e., Test-(Deceiving Server)) and the target data (that is, Target-(Deceiving Server)) is almost the same, when the server honestly executes unlearning. However, if the server unlearns the irrelevant data to replace the target data, Target-(Deceiving Server) will be significantly reduced. 
Because the information of target data is still preserved in the unlearned model $M_u$, they do not cause greater gradient changes to the model than the test data.

\section{\name in Face Recognition} \label{subsec:UMI-face}

Here, we apply \name in the face recognition system, which is often deployed in real-world, e.g., by the company and community. When the data contributor resigns or moves, he/she would request the company and community to remove his/her facial information. As we mentioned above, machine unlearning is non-trivial, so the model owner perhaps ignores the unlearning request (i.e., the Neglecting Server). For such a situation, we verify the unlearning result by \name.
\begin{figure}
	\centering
	\subfigure[Unlearning-Metric-I]{
		\includegraphics[width = 0.387\textwidth]{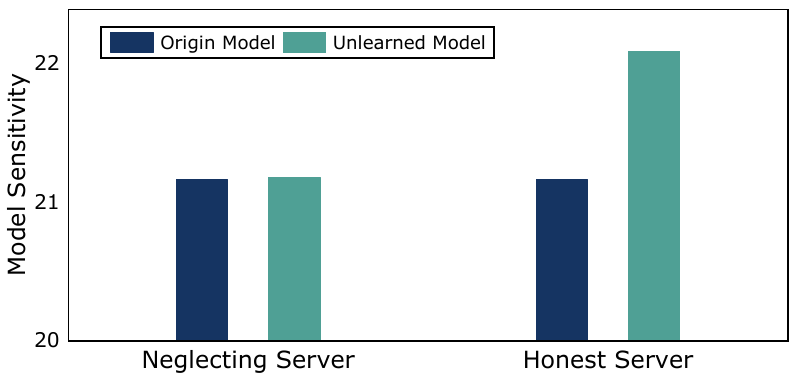}
		
	}
	\subfigure[Unlearning-Metric-III]{
		\includegraphics[width = 0.4\textwidth]{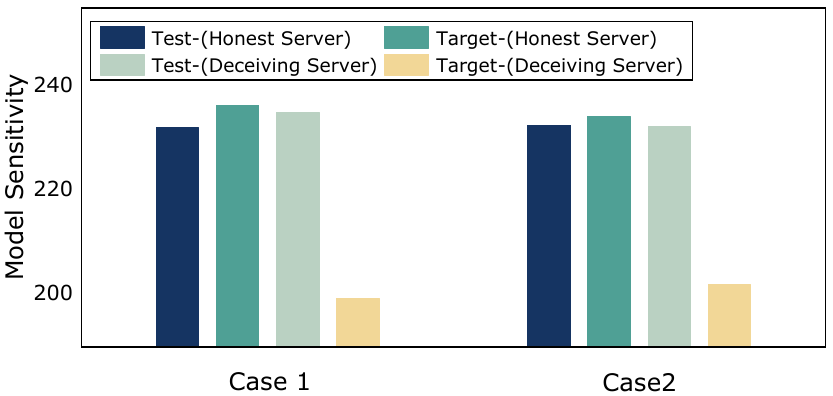}
	}
	\caption{\name's performance in the face recognition system.}
	\label{Fig14:UM-I-face}
\end{figure}

In the experiment, we train the model on the Pubfig dataset \cite{KumarBBN09}, which is a large, real-world face dataset consisting of 58,797 images of 200 people collected from the internet. We use the VGG16 to train the model, and select 68 people randomly with 1 person we choose to unlearn. The Neglecting Server could be trivially captured in the performance of Unlearning-Metric-I, see Figure \ref{Fig14:UM-I-face} (a). The dishonest behavior is depicted clearly in the difference in model sensitivity between origin model and unlearned model: honest unlearning could raise the model sensitivity of the target class, but neglecting unlearning could not. And the deviation of Unlearning-Metric-II is 8.6\%, which could identify the Lazy Server. The verification result of Unlearning-Metric-III is shown in Figure \ref{Fig14:UM-I-face} (b). The Honest Server and Deceiving Server exhibit a significant discrepancy in model sensitivity between train data and test data under Unlearning-Metric-III, thereby demonstrating the effective validation capability of \name regarding the presence of target data.

\section{Non-IID Model}
To demonstrate the effectiveness of \name given a Non-IID dataset, we conducted experiments of \name on the GTSRB dataset \cite{StallkampSSI11}, with 39,209 training images and 12,630 testing images of 43 different classes---samples per each class is imbalanced for this dataset. 
We randomly select 10 classes, and unlearn all their samples. Then we evaluate \name, and the verification accuracy of Unlearning-Metric-I is  0.91, the verification deviation of Unlearning-Metric-II is 5.8\% per category on average, and the verification accuracy of Unlearning-Metric-III is 0.88. The experimental results demonstrate that \name is independent of the data distribution of the subject dataset.

\section{Algorithms} \label{sec:algorithm}

The detailed algorithms of three unlearning metrics in \name are shown below:

\begin{algorithm}[htbp]
	\caption{Unlearned Class Verification}
	\label{alg2:Unlearning-Metric-I}
	\LinesNumbered 
	\KwIn{\textbf{ The origin model $\theta_{o}$, the unlearned model $\theta_{u}$,  test data of target class $D_{\rm{test}}$, learning rate $\alpha$}}
	\KwOut{\textbf{Verification Result $result$}, \textbf{Unlearned Class $c$}}
	Extract model sensitivity of $\theta_{o}$: $MS_{o}$ =$\theta _{o}$. \textsf{Extract}($D_{\rm{test}}$,$\alpha$  ) \;
	Extract model sensitivity of $\theta_{u}$: $MS_{u}$ =$\theta _{u}$. \textsf{Extract}($D_{\rm{test}}$,$\alpha$  ) \;
	Match each class $c$ in ($MS_{o}$, $MS_{u}$) \;
		\eIf{$MS_{o}^{c}$$\approx$$MS_{u}^{c}$ }{
			$result$=$false$, which represents the server \textbf{does not} perform any unlearning operation on the target data\;
		}{
			$result$=$true$, which represents the server indeed unlearned the data of class $c$\;
		}
\end{algorithm}

\begin{algorithm}[htbp]
	\caption{Unlearned Volume Verification}
	\label{alg3:Unlearning-Metric-II}
	\LinesNumbered 
	\KwIn{\textbf{ The origin model $\theta_{o}$, the unlearned model $\theta_{u}$,  test data of target class $D_{\rm{test}}$, target data $D_{\rm{tar}}$, learning rate $\alpha$}}
	\KwOut{\textbf{Estimated volume of unlearned samples $Target~data~volume$}}
	Extract model sensitivity of $\theta_{o}$: $MS_{o}$ =$\theta _{o}$. \textsf{Extract}($D_{\rm{test}}$,$\alpha$  ) \;
	Extract model sensitivity of $\theta_{u}$: $MS_{u}$ =$\theta _{u}$. \textsf{Extract}($D_{\rm{test}}$,$\alpha$  ) \;
	Compute $DS$=$MS_{u}$-$MS_{o}$\;
	\textbf{Get the unlearning measurement:}\\
	Divide $D_{\rm{tar}}^{c}$ of target class $c$ into $n$ slides incrementally: $D_{\rm{tar}}^{c}$ $\overset{}{\rightarrow}$ $\left \{ D_{\rm{shadow_{1} }}^{c},~D_{\rm{shadow_{2}}}^{c},~...~, D_{\rm{shadow_{n} }}^{c}\right\}$, and record the increasing number of shadow datasets $batch~volume$\;
	Train $n$ shadow models $\theta_{\rm{shadow_n}}$ based on $D_{\rm{shadow_{n}}}$\;
	Extract model sensitivity of $n$ shadow models: $MS_{\rm{shadow_n}}$ =$\theta _{\rm{shadow_n}}$. \textsf{Extract}($D_{\rm{test}}$,$\alpha$) \;
	Compute the model sensitivity difference: $DS'_{n}=MS_{\rm{shadow_{n+1}}}- MS_{\rm{shadow_n}}$\;
	Generate unlearning measurement: $UM_{\rm{batch}}=\frac{\sum_{i=1}^{n}DS'_{\rm{2i-1}}}{n}$\;
	Estimate volume of unlearned samples: $Target~data~volume=\left \lceil \frac{DS}{UM_{\rm{batch}} }  \right \rceil \times batch~volume$.
\end{algorithm}

\begin{algorithm} [htbp]
	\caption{Unlearned Sample Verification}
	\label{alg4:Unlearning-Metric-III}
	\LinesNumbered 
	\KwIn{\textbf{ The unlearned model $\theta_{u}$, target data $D_{\rm{tar}}$, test data $D_{\rm{test}}$, learning rate $\alpha$}}
	\KwOut{\textbf{Verification Result $result$}}
	Extract model sensitivity of $\theta_{u}$ based on $D_{\rm{test}}$: $MS_{\rm{u\_test}}$ =$\theta _{u}$. \textsf{Extract}($D_{\rm{test}}$,$\alpha$  ) \;
	Extract model sensitivity of $\theta_{u}$ based on $D_{\rm{tar}}$: $MS_{\rm{u\_tar}}$ =$\theta _{u}$. \textsf{Extract}($D_{\rm{tar}}$,$\alpha$  ) \;
	Model sensitivity similarity check with ($MS_{\rm{u\_test}}$, $MS_{\rm{u\_tar}}$)\; 
		\eIf{$MS_{\rm{u\_test}}$$\approx$$MS_{\rm{u\_tar}}$ }{
			$result$=$true$, which represents the server indeed unlearns the target sample \;
		}{
			$result$=$false$, which represents the target sample \textbf{still remains} in the unlearned model, and the server has spoofed the data provider and data contributor.\;
		}
\end{algorithm}

\end{document}